\documentclass[10pt,twocolumn,letterpaper]{article}

\usepackage{times}
\usepackage{epsfig}
\usepackage{graphicx}
\usepackage{amsmath}
\usepackage{amssymb}
\usepackage{sidecap}
\usepackage{epstopdf}
\usepackage{multirow}

\usepackage{caption}
\usepackage{subcaption}

\newsavebox\CBox

\newcommand{\eg}{{\emph{e.g.}}}
\newcommand{\ie}{{\emph{i.e.}}}
\newcommand{\etal}{{\emph{et al.\ }}}
\newcommand{\wrt}{{\emph{w.r.t.}\ }}

\usepackage[pagebackref=true,breaklinks=true,letterpaper=true,colorlinks,bookmarks=false]{hyperref}
\usepackage[top = 2.00cm, bottom = 2.00cm, left = 2.0cm, right = 2.25cm]{geometry}

\begin{document}


\title{WESPE: Weakly Supervised Photo Enhancer for Digital Cameras}

\author{Andrey Ignatov\\
{\tt\small andrey.ignatoff@gmail.com}
\and
Nikolay Kobyshev\\
{\tt\small nk@vision.ee.ethz.ch}
\and
Kenneth Vanhoey\\
{\tt\small vanhoey@vision.ee.ethz.ch}
\and
Radu Timofte\\
{\tt\small timofter@vision.ee.ethz.ch}
\and
Luc Van Gool\\
{\tt\small vangool@vision.ee.ethz.ch}
\\
}

\author{Andrey Ignatov, \hspace{2mm} Nikolay Kobyshev,  \hspace{2mm} Kenneth Vanhoey,  \hspace{2mm} Radu Timofte,  \hspace{2mm} Luc Van Gool \\
\\
\textsf{ETH Zurich}\\
\\
{\tt\small \{andrey, nk, vanhoey, timofter, vangool\}@vision.ee.ethz.ch}\\
}

\date{}

\maketitle

\begin{abstract}
Low-end and compact mobile cameras demonstrate limited photo quality mainly due to space, hardware and budget constraints. In this work, we propose a deep learning solution that translates photos taken by cameras with limited capabilities into DSLR-quality photos automatically. We tackle this problem by introducing a weakly supervised photo enhancer (WESPE) -- a novel image-to-image Generative Adversarial Network-based architecture. The proposed model is trained by under weak supervision: unlike previous works, there is no need for strong supervision in the form of a large annotated dataset of aligned original/enhanced photo pairs. The sole requirement is two distinct datasets: one from the source camera, and one composed of arbitrary high-quality images that can be generally crawled from the Internet~-- the visual content they exhibit may be unrelated.
Hence, our solution is repeatable for any camera: collecting the data and training can be achieved in a couple of hours.
In this work, we emphasize on extensive evaluation of obtained results.
Besides standard objective metrics and subjective user study, we train a virtual rater in the form of a separate CNN that mimics human raters on Flickr data and use this network to get reference scores for both original and enhanced photos. Our experiments on the DPED, KITTI and Cityscapes datasets as well as pictures from several generations of smartphones demonstrate that WESPE produces comparable or improved qualitative results with state-of-the-art strongly supervised methods.
\end{abstract}

\begin{figure}[th!]
\centering
\begin{tabular}{c}
\includegraphics[width=0.84\linewidth]{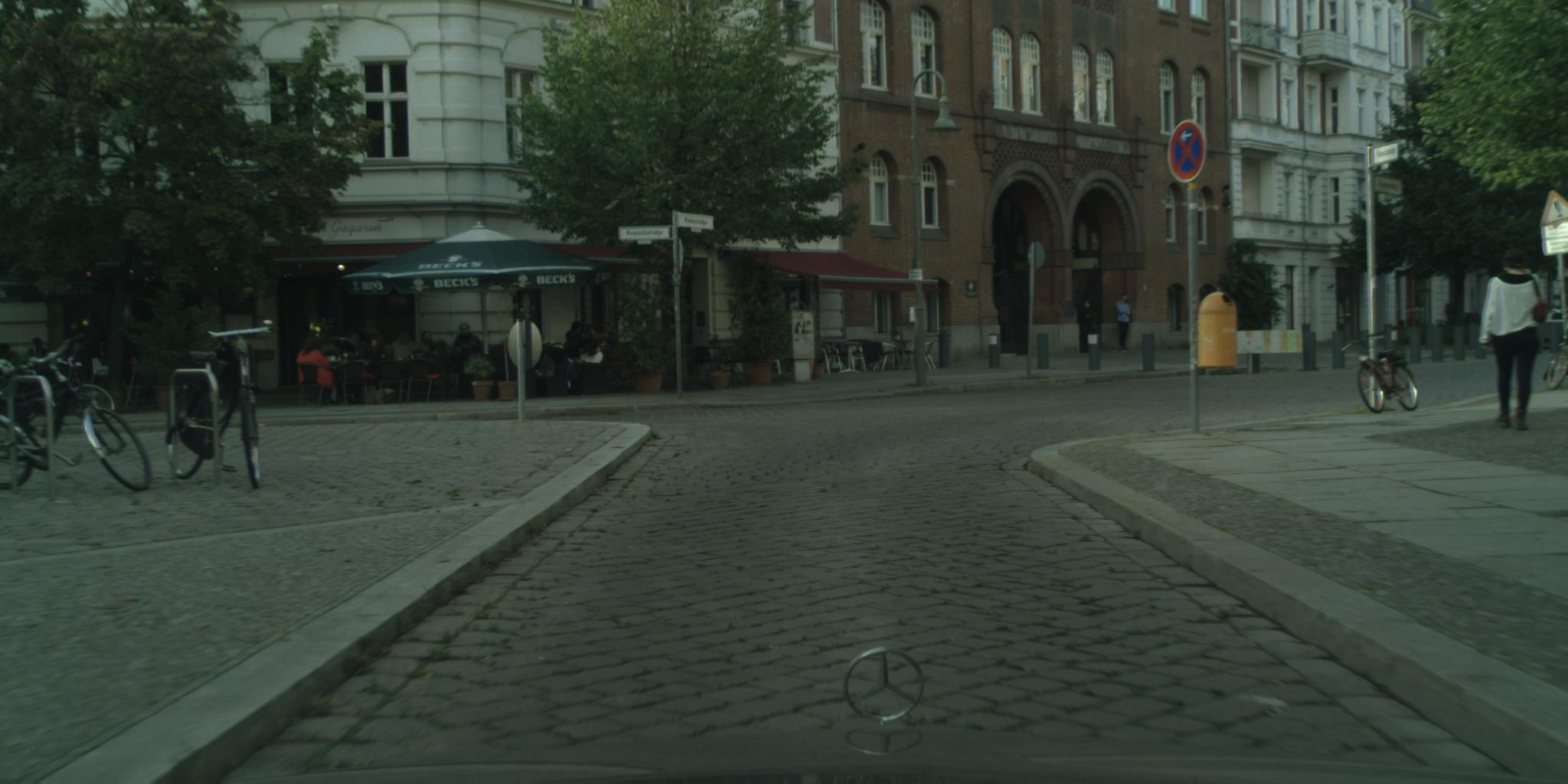}\\
\, \, $\blacktriangledown$ \vspace{0.8mm} \\
\includegraphics[width=0.84\linewidth]{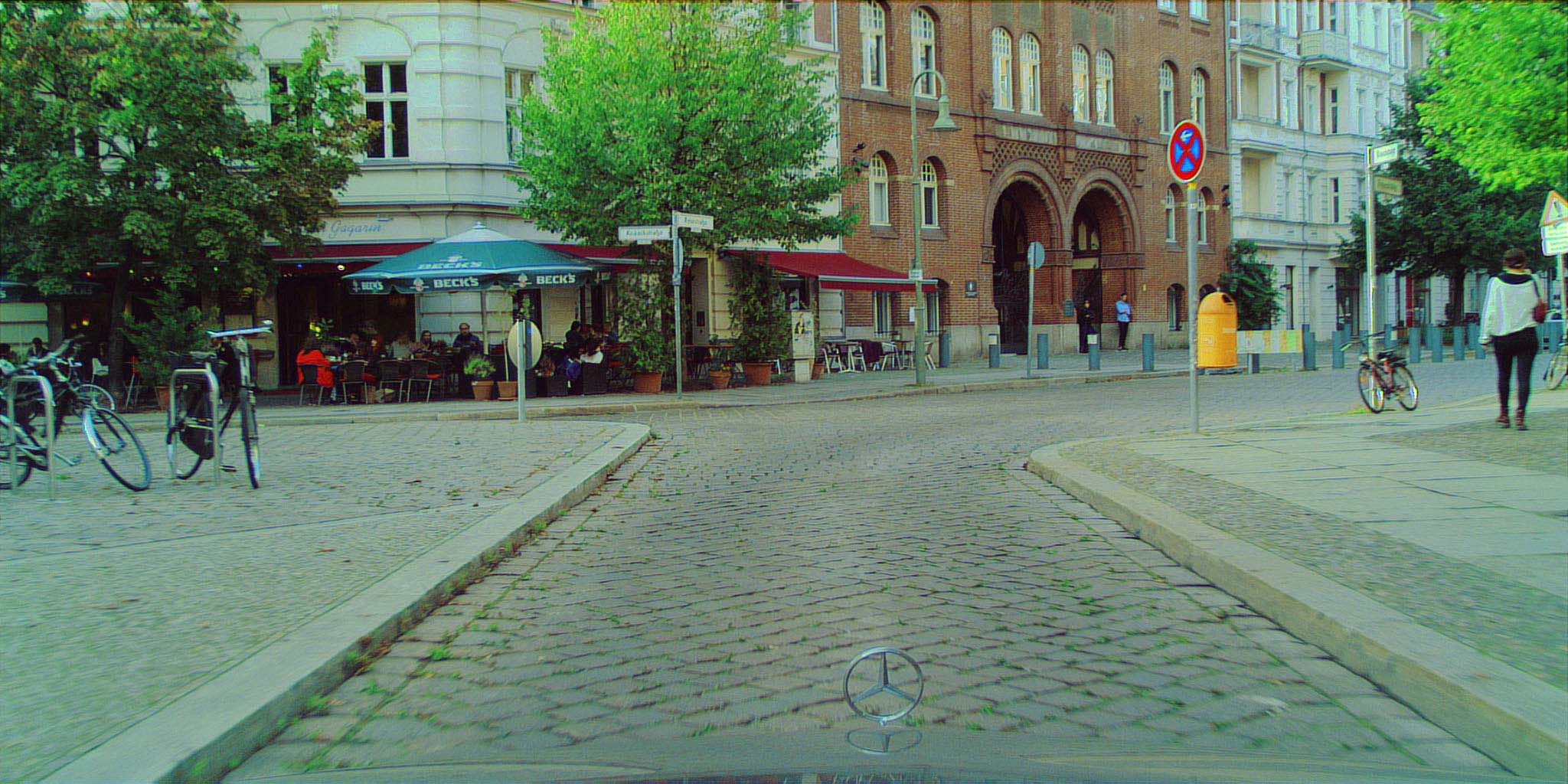}
\end{tabular}
\caption{Cityscapes image enhanced by our method.}
\label{fig:short}
\end{figure}

\section{Introduction}
\label{sec:introduction}
The ever-increasing quality of camera sensors allows us to photograph scenes with unprecedented detail and color.
But as one gets used to better quality standards, photos captured just a few years ago with older hardware look dull and outdated.
Analogously, despite incredible advancement in quality of images captured by mobile devices, compact sensors and lenses make DSLR-quality unattainable for them, leaving casual users with a constant dilemma of relying on their lightweight mobile device or transporting a heavier-weight camera around on a daily basis.
However, the second option may not even be possible for a number of other applications such as autonomous driving or video surveillance systems, where primitive cameras are usually employed.

In general, image enhancement can be done manually (\eg, by a graphical artist) or semi-automatically using specialized software capable of histogram equalization, photo sharpening, contrast adjustment, etc. The quality of the result in this case significantly depends on user skills and allocated time, and thus is not doable by non-graphical experts on a daily basis, or not applicable in case of real-time or large-scale data processing.
A fundamentally different option is to train various learning-based methods that allow to automatically transform image style or to perform image enhancement. Yet, one of the major bottlenecks of these solutions is the need for strong supervision using matched before/after training pairs of images.
This requirement is often the source of a strong limitation of color/texture transfer~\cite{OVBED15} and photo enhancement~\cite{IKTVvG17} methods.

In this paper, we present a novel weakly supervised solution for the image enhancement problem to deliver ourselves from the above constraints. That is, we propose a deep learning architecture that can be trained to enhance images by mapping them from the domain of a given source camera into the domain of high-quality photos (supposedly taken by high-end DSLRs) while not requiring any correspondence or relation between the images from these domains: only two separate photo collections representing these domains are needed for training the network. To achieve this, we take advantage of two novel advancements in Generative Adversarial Networks (GAN) and Convolutional Neural Networks (CNN): \textbf{i)} transitive CNNs to map the enhanced image back to the space of source images so as to relax the need of paired ground truth photos~\cite{zhu2017unpaired}, and \textbf{ii)} loss functions combining color, content and texture loss to learn photorealistic image quality~\cite{IKTVvG17}.
The key advantage of the method is that it can be learned easily: the training data is trivial to obtain for any camera and training takes just a few hours.
Yet, quality-wise, our results still surpass traditional enhancers and compete with state of the art (fully supervised) methods by producing artifact-less results.

\smallskip

\textbf{Contributions.}
Enhanced images improve the non-enhanced ones in several aspects, including colorization, resolution and sharpness.
Our contributions include:
\vspace{-0.7mm}
\begin{itemize}
\item[$\bullet$] \textit{WESPE}, a generic method for learning a model that enhances source images into DSLR-quality ones,
\vspace{-1.2mm}
\item[$\bullet$] a transitive CNN-GAN architecture, made suitable for the task of image enhancement and image domain transfer by combining state of the art losses with a content loss expressed on the input image,
\vspace{-1.2mm}
\item[$\bullet$] large-scale experiments on several publicly available datasets with a variety of camera types, including subjective rating and comparison to the state of the art enhancement methods,
\vspace{-1.2mm}
\item[$\bullet$] a \textit{Flickr Faves Score} (FFS) dataset consisting of 16K HD resolution Flickr photos with an associated number of likes and views that we use for training a separate scoring CNN to independently assess image quality of the photos throughout our experiments,
\vspace{-1.2mm}
\item[$\bullet$] openly available models and code\footnote{\url{http://people.ee.ethz.ch/~ihnatova/wespe.html}}, that we progressively augment with additional camera models / types.
\end{itemize}

\section{Related work}
\label{sec:related_work}
Automatic photo enhancement can be considered as a typical --~if not the ultimate~-- computational photography task.
To devise our solution, we build upon three sub-fields: style transfer, image restoration and general-purpose image-to-image enhancers.

\subsection{Style transfer}
The goal of style transfer is to apply the style of one image to the (visual) content of another.
Traditional texture/color/style transfer techniques~\cite{EF01,HJOCS01,liu2014autostyle,OVBED15} rely on an exemplar before/after pair that defines the transfer to be applied.
The exemplar pair should contain visual content having a sufficient level of analogy to the target image's content
which is hard to find, and this hinders its automatic and mass usage.
More recently, neural style transfer alleviates this requirement~\cite{GEB15,UlyanovLVL16}.
It builds on the assumption that the shallower layers of a deep CNN classifier -- or more precisely, their correlations -- characterize the style of an image, while the deeper ones represent semantic content.
A neural network is then used to obtain an image matching the style of one input and the content of another.
Finally, generative adversarial networks (GAN) append a discriminator CNN to a generator network~\cite{GAN}.
The role of the former is to distinguish between two domains of images: \eg, those having the style of the target image and those produced by the generator.
It is jointly trained with the generator, whose role is in turn to fool the discriminator by generating an image in the right domain, \ie, the domain of images of correct style.
We exploit this logic to force the produced images to be in the domain of target high-quality photos.

\subsection{Image restoration}
Image quality enhancement has traditionally been addressed through a list of its sub-tasks, like super-resolution, deblurring, dehazing, denoising, colorization and image adjustment.
Our goal of hallucinating high-end images from low-end ones encompasses all these enhancements.
Many of these tasks have recently seen the arrival of successful methods driven by deep learning phrased as image-to-image translation problems.
However, a common property of these works is that they are targeted at \emph{removing artifacts added artificially} to clean images, thus requiring to model all possible distortions.
Reproducing the flaws of the optics of one camera compared to a high-end reference one is close to impossible, let alone repeating this for a large list of camera pairs.
Nevertheless, many useful ideas have emerged in these works, their brief review is given below.

The goal of \emph{image super-resolution} is to restore the original image from its downscaled version.
Many end-to-end CNN-based solutions exist now~\cite{Dong2014,Kim2016,Shi2016,Mao2016,timofte2017ntire}.
Initial methods used pixel-wise mean-squared-error (MSE) loss functions, which often generated blurry results.
Losses based on the activations of (a number of) VGG-layers~\cite{Johnson2016} and GANs~\cite{Ledig2016} are more capable of recovering photorealistic results, including high-frequency components, hence produce state of the art results.
In our work, we incorporate both the GAN architectures and VGG-based loss functions.

\emph{Image colorization}~\cite{Richard2016,Cheng2015,luan2007natural}, which attempts to regress the 3 RGB channels from images that were reduced to single-channel grayscale, strongly benefits from the GAN architecture too~\cite{Isola_2017_CVPR}.
Image \emph{denoising}, \emph{deblurring} and \emph{dehazing}~\cite{Zhang2016,Svoboda2016,Hradis2015,Ling2016,Cai2016}, photographic \emph{style control}~\cite{Yan2016} and \emph{transfer}~\cite{Lee2016}, as well as \emph{exposure correction}~\cite{Yuan2012} are another improvements and adjustments that are included in our learned model.
As opposed to mentioned related work, there is no need to manually model these effects in our case.

\subsection{\hspace{-2mm}\mbox{General-purpose image-to-image enhancers}}
We build our solution upon very recent advances in image-to-image translation networks.
Isola~\etal~\cite{Isola_2017_CVPR} present a general-purpose translator that takes advantage of GANs to learn the loss function depending on the domain the target image should be in.
While it achieves promising results when transferring between very different domains (\eg, aerial image to street map), it lacks photorealism when generating photos: results are often blurry and with strong checkerboard artifacts.
Compared to our work, it needs \emph{strong supervision}, in the form of many before/after examples provided at training time.

Zhu~\etal~\cite{zhu2017unpaired} loosen this constraint by expressing the loss in the space of input rather than output images, taking advantage of a backward mapping CNN that transforms the output back into the space of input images.
We apply a similar idea in this work.
However, our CNN architecture and loss functions are based on different ideas: fully convolutional networks and elaborated losses allow us to achieve photorealistic results, while eliminating typical artifacts (like blur and checkerboard) and limitations of encoder-decoder networks.

Finally, Ignatov~\etal~\cite{IKTVvG17} propose an end-to-end enhancer achieving photorealistic results for arbitrary-sized images due to a composition of content, texture and color losses.
However, it is trained with a strong supervision requirement for which a dataset of aligned ground truth image pairs taken by different cameras was assembled (\ie, the DPED dataset).
We build upon their loss functions to achieve photorealism as well, while adapting them to the new architecture suitable for our weakly supervised learning setting.
While we do not need a ground truth aligned dataset, we use DPED to report the performance on. Additionally, we provide the results on public datasets (KITTI, Cityscapes) and several newly collected datasets for smartphone cameras.

\begin{figure}
\centering
 \includegraphics[width=\linewidth]{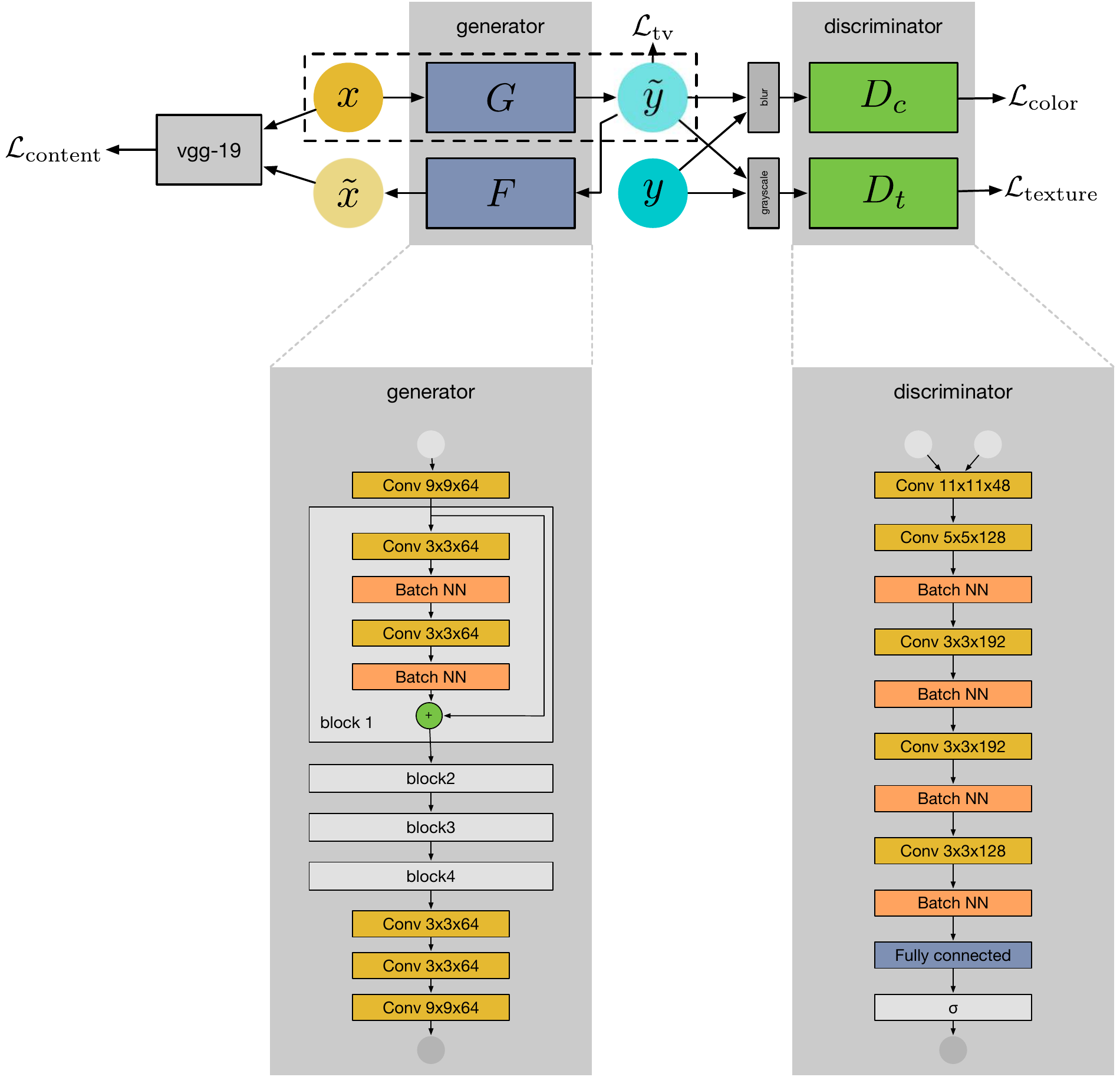}
\caption{Proposed WESPE architecture.}
\label{fig:proposed_method}
\end{figure}

\section{Proposed method}
\label{sec:proposed_method}
Our goal is to learn a mapping from a source domain $X$ (\eg, defined by a low-end digital camera) to a target domain $Y$ (\eg, defined by a collection of captured or crawled high-quality images).
The inputs are unpaired training image samples ${x}\in X$ and ${y}\in Y$.
As illustrated in Figure~\ref{fig:proposed_method}, our model consists of a generative mapping $G:X\rightarrow Y$ paired with an inverse generative mapping $F:Y\rightarrow X$.
To measure content consistency between the mapping $G(x)$ and the input image $x$, a content loss based on VGG-19 features is defined between the original and reconstructed images $x$ and $\tilde{x}=(F \circ G) (x)$, respectively.
Defining the \textit{content loss} in the input image domain allows us to circumvent the need of before/after training pairs.
Two adversarial discriminators $D_c$ and $D_t$ and total variation (TV) complete our loss definition.
$D_c$ aims to distinguish between high-quality image $y$ and enhanced image $\tilde{y}=G(x)$ based on image colors, and $D_t$ based on image texture.
As a result, our objective comprises:
i) content consistency loss to ensure $G$ preserves $x$'s content,
ii) two adversarial losses ensuring generated images $\tilde{y}$ lie in the target domain $Y$: a color loss and a texture loss,
and iii) TV loss to regularize towards smoother results.
We now detail each of these loss terms.

\vspace{-2.8mm}

\paragraph{3.1. Content consistency loss.}
\label{ssc:content_loss}

We define the \emph{content consistency loss} in the input image domain~$X$: that is, on $x$ and its reconstruction $\tilde x = F(\tilde{y}) = F\circ G(x)$ (inverse mapping from the enhanced image), as shown in Figure~\ref{fig:proposed_method}.
Our network is trained for both the direct $G$ and inverse $F$ mapping simultaneously, aiming at strong content similarity between the original and enhanced image.
We found pixel-level losses too restrictive in this case, hence we choose a perceptual content loss based on ReLu activations of the VGG-19 network~\cite{simonyan2014very}, inspired by~\cite{IKTVvG17,Johnson2016,Ledig2016}.
It is defined as the $l_2$-norm between feature representations of the input image $x$ and the recovered image $\tilde x$:
\begin{equation}
\mathcal{L}_{\text{content}} = \frac{1}{C_jH_jW_j} \|\psi_j\bigl(x\bigr) - \psi_j\bigl(\tilde{x}\bigr)\|,
\end{equation}
where $\psi_j$ is the feature map from the $j$-th VGG-19 convolutional layer and $C_j$, $H_j$ and $W_j$ are the number, height and width of the feature maps, respectively.

\vspace{-2.8mm}

\paragraph{3.2. Adversarial color loss.}
\label{ssc:color_loss}

Image color quality is measured using an adversarial discriminator $D_c$ that is trained to differentiate between the blurred versions of enhanced $\tilde{y}_b$ and high-quality $y_b$ images:
\begin{equation}
y_b(i,j) = \sum_{k,l} y(i+k,j+l) \cdot G_{k,l}, \\
\end{equation}
where
$
G_{k,l} = A \exp\bigl(-\frac{(k - \mu_x)^2}{2\sigma_x} -\frac{(l - \mu_y)^2}{2\sigma_y}\bigr)
$
defines Gaussian blur with $A=0.053$, $\mu_{x,y}=0$, and $\sigma_{x,y}=3$ set empirically.

The main idea here is that the discriminator should learn the differences in brightness, contrast and major colors between low-- and high-quality images, while it should avoid texture and content comparison.
A constant $\sigma$ was defined experimentally to be the smallest value that ensures texture and content eliminations.
The loss itself is defined as a standard generator objective, as used in GAN training:
\begin{equation}
\mathcal{L}_{\text{color}} = -\sum_{i} \log{D_c(G(x)_b)}.
\end{equation}
Thus, color loss forces the enhanced images to have similar color distributions as the target high-quality pictures.

\vspace{-2.8mm}

\paragraph{3.3. Adversarial texture loss.}
\label{ssc:texture_loss}

Similarly to color, image texture quality is also assessed by an adversarial discriminator $D_t$.
This is applied to grayscale images and is trained to predict whether a given image was artificially enhanced~($\tilde{y}_g$) or is a ``true'' native high-quality image ($y_g$).
As in the previous case, the network is trained to minimize the cross-entropy loss function, the loss is defined as:
\begin{equation}
\mathcal{L}_{\text{texture}} = -\sum_{i} \log{D_t(G(x)_g)}.
\end{equation}
As a result, minimizing this loss will push the generator to produce images of the domain of native high-quality ones.

\vspace{-2.8mm}

\paragraph{3.4. TV loss.}
\label{ssc:TV_loss}
To impose spatial smoothness of the generated images we also add a total variation loss~\cite{TV2005} defined as follows:
\vspace{-2mm}
\begin{equation}
\mathcal{L}_{\text{tv}} = \frac{1}{C H W} \|\nabla_x G(x) + \nabla_y G(x)\|,
\end{equation}
where $C$, $H$, $W$ are dimensions of the generated image $G(x)$.

\vspace{-2.8mm}

\paragraph{3.5. Sum of losses.}
The final WESPE loss is composed of a linear combination of the four aforementioned losses:
\begin{equation}
\mathcal{L}_{\text{total}} = \mathcal{L}_{\text{content}} \ + \ 5\cdot10^{-3} \ (\mathcal{L}_{\text{color}} + \mathcal{L}_{\text{texture}}) + 10\ \mathcal{L}_{\text{tv}}.
\end{equation}
The weights were picked based on preliminary experiments on our training data.

\vspace{-2.8mm}

\paragraph{3.6. Network architecture and training details.}

The overall architecture of the system is illustrated in Figure~\ref{fig:proposed_method}. Both generative and inverse generative networks $G$ and $F$ are fully-convolutional residual CNNs with four residual blocks, their architecture was adapted from~\cite{IKTVvG17}. The discriminator CNNs consist of five convolutional and one fully-connected layer with 1024 neurons, followed by the last layer with a sigmoid activation function on top of it. The first, second and fifth convolutional layers are strided with a step size of 4, 2 and 2, respectively. For each dataset the train/test splits are as shown in Tables~\ref{tab:dped_dataset} and~\ref{tab:in_the_wild_datasets}.

The network was trained on an \textit{NVIDIA Titan X} GPU for 20K iterations using a batch size of 30 and the size of the input patches was 100$\times$100 pixels. The parameters of the networks were optimized using the Adam algorithm. The experimental setup was identical in all experiments.


\begin{figure*}[t!]
\centering
\setlength{\tabcolsep}{1pt}
\resizebox{\linewidth}{!}
{
\begin{tabular}{ccc}
   \includegraphics[width=0.33\linewidth]{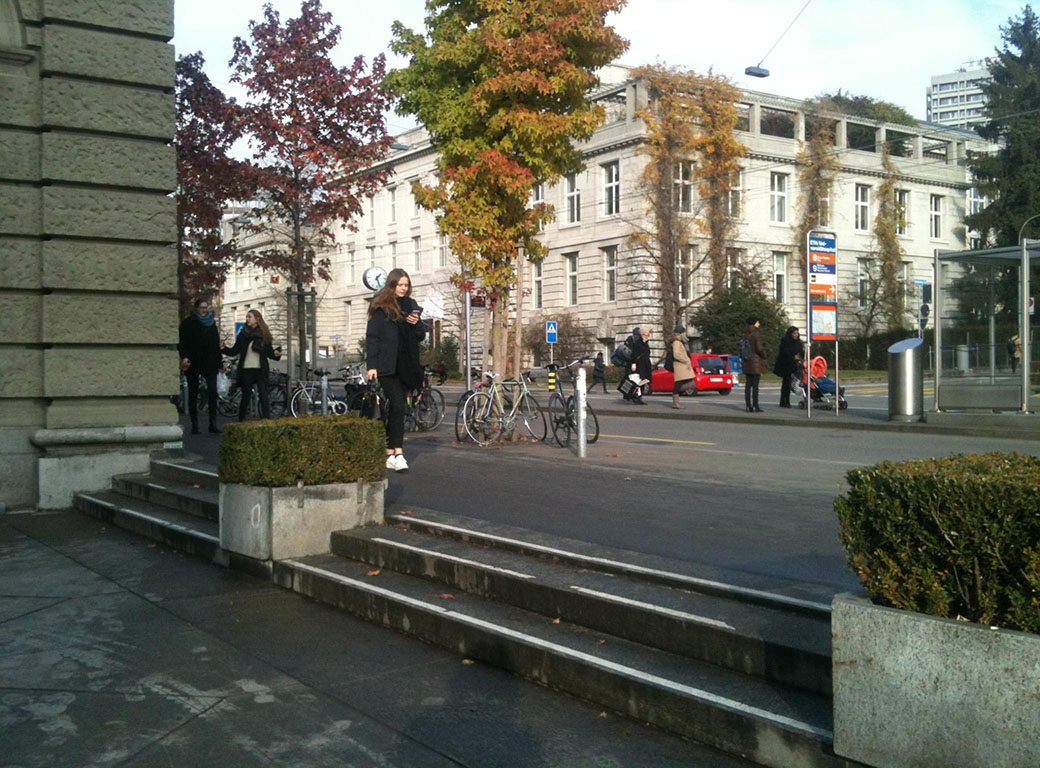}&
   \includegraphics[width=0.33\linewidth]{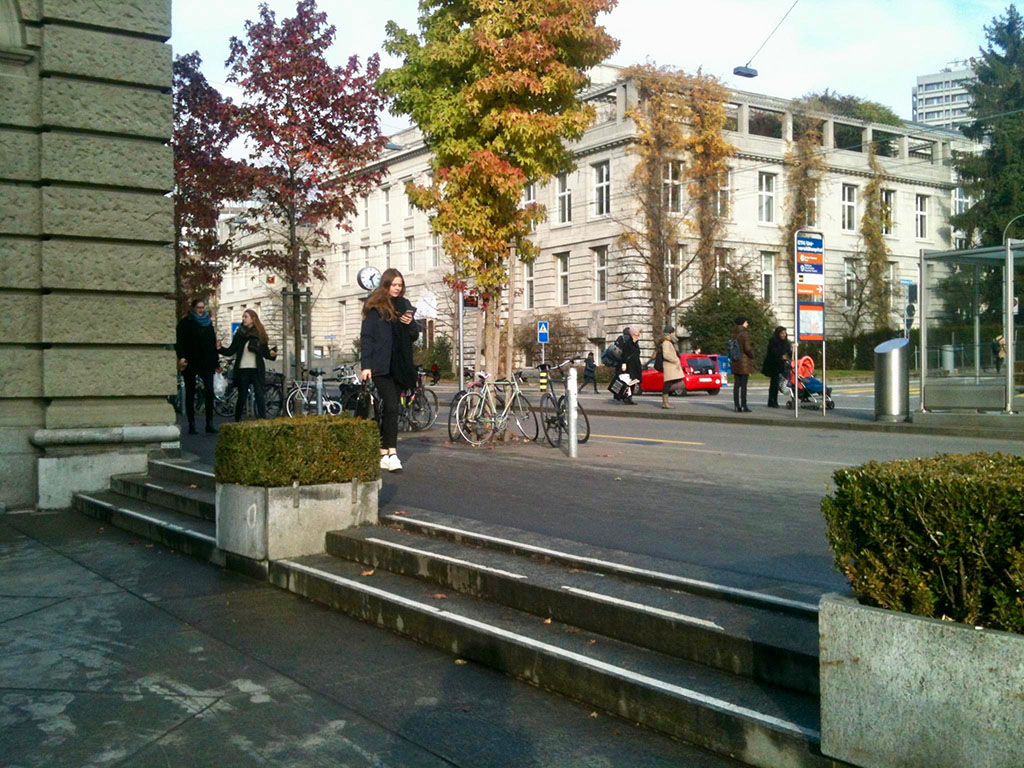}&
   \includegraphics[width=0.33\linewidth]{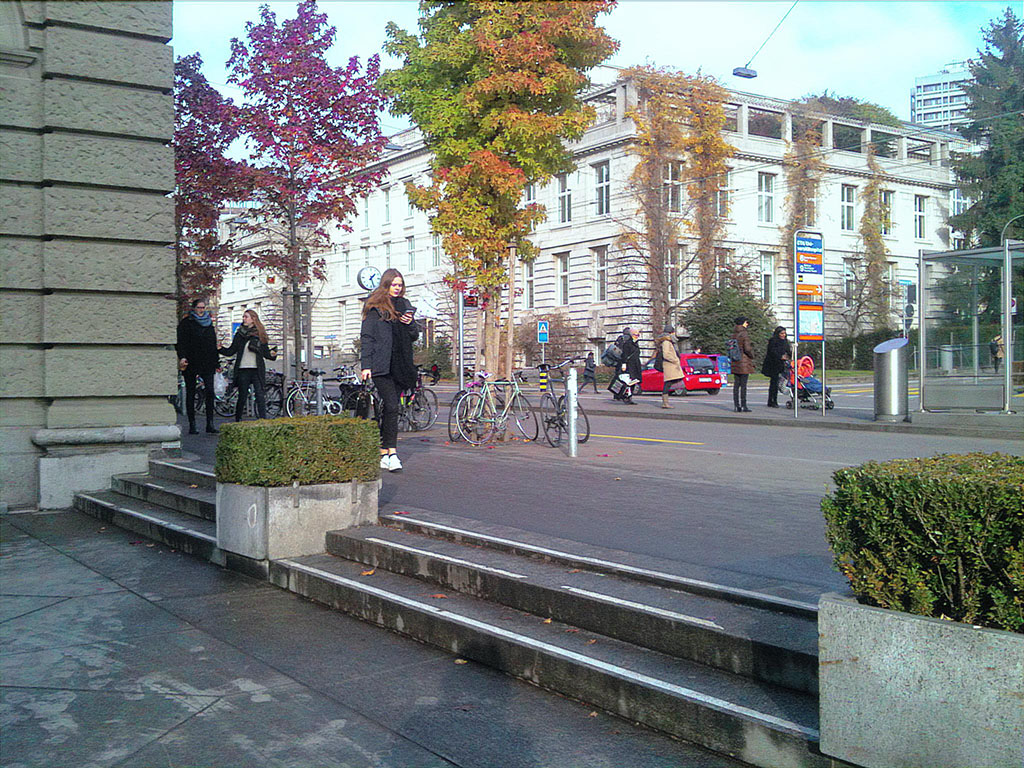}\\

   \vspace{0.5mm}

   \includegraphics[width=0.33\linewidth]{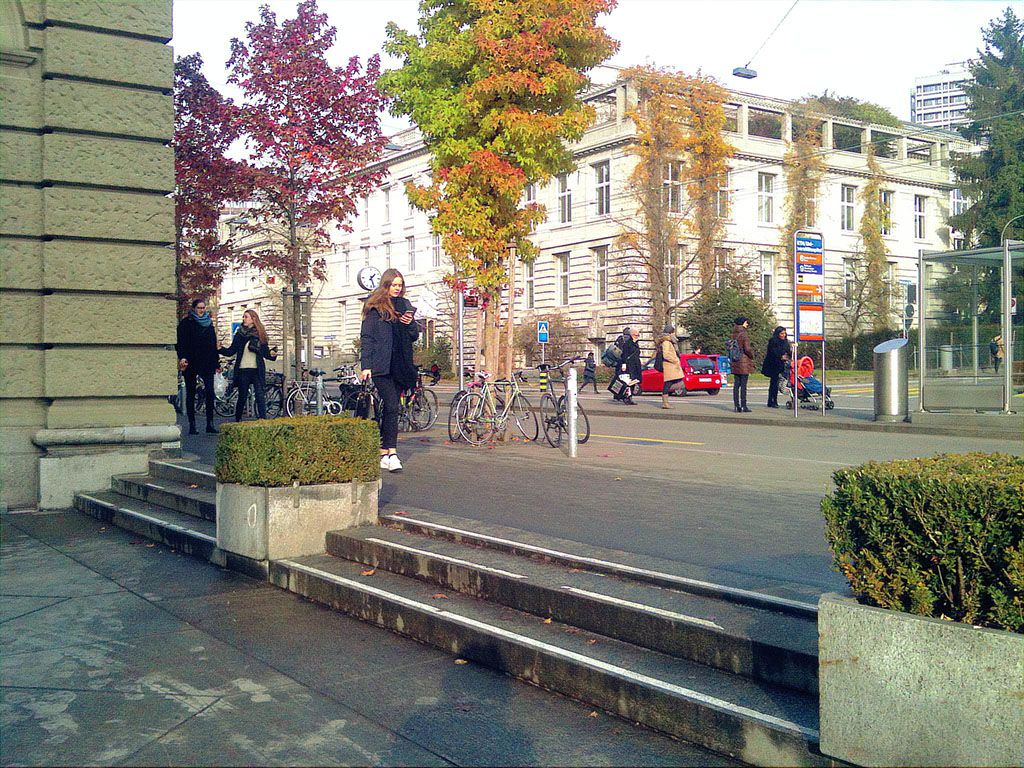}&
   \includegraphics[width=0.33\linewidth]{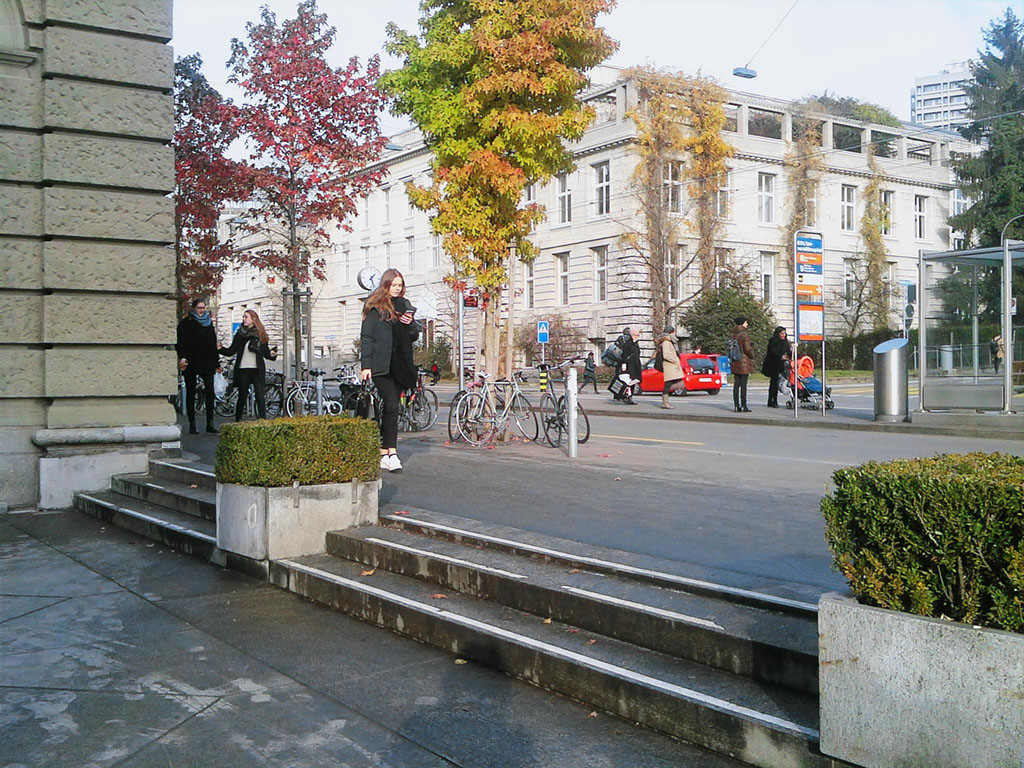}&
   \includegraphics[width=0.33\linewidth]{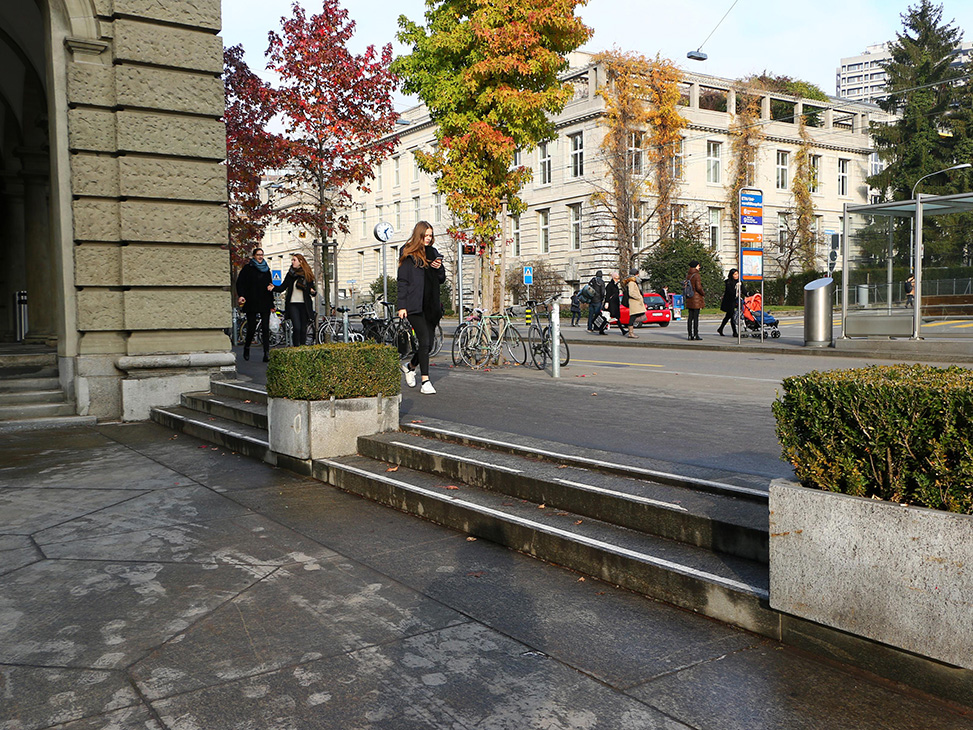}\\
   \end{tabular}
}
\vspace{-2mm}
 \caption{From left to right, top to bottom: original iPhone photo and the same image after applying, resp.: Apple Photo Enhancer, WESPE trained on DPED, WESPE trained on DIV2K, Ignatov~\etal~\cite{IKTVvG17}, and the corresponding DSLR image.}
 \vspace{-2mm}
 \label{fig:all_enhancements1}
\end{figure*}

\section{Experiments}
\label{sec:experiments}

To assess the abilities and quality of the proposed network (WESPE), we apply a series of experiments covering several cameras and datasets.
We also compare against a commercial software baseline (the Apple Photos image enhancement software, or APE, version 2.0) and the latest state of the art in the field by Ignatov~\etal~\cite{IKTVvG17}, that uses learning under full supervision.
We start our experiments by doing a full-reference quantitative evaluation of the proposed approach in section~\ref{sec:experiments:dped}, using the ground truth DPED dataset used for supervised training by Ignatov~\etal~\cite{IKTVvG17}.
WESPE however is unsupervised, so it can be applied to any dataset in the wild as no ground truth enhanced image is needed for training.
In section~\ref{sec:experiments:wespe} we apply WESPE on such datasets of various nature and visual quality, and evaluate quantitatively using no-reference quality metrics.
Since the main goal of WESPE is qualitative performance which is not always reflected by conventional metrics, we additionally use subjective evaluation of the obtained results.
Section~\ref{sec:experiments:userstudy} presents a study involving human raters, and in section~\ref{sec:experiments:ffs} we build and use a Flickr faves score emulator to emulate human rating on a large scale.
For all experiments, we also provide qualitative visual results.

\subsection{Full-reference evaluation}
\label{sec:experiments:dped}

In this section, we perform our experiments on the the DPED dataset (see Table~\ref{tab:dped_dataset}) that was initially proposed for learning a photo enhancer with full supervision~\cite{IKTVvG17}.
DPED is composed of images from three smartphones with low --to middle-end cameras (\ie, iPhone 3GS, BlackBerry Passport and Sony Xperia Z) paired with images of the same scenes taken by a high-end DSLR camera (\ie, Canon 70D) with pixel alignment.
Thanks to this pixel-aligned ground truth before/after data, we can exploit full-reference image quality metrics to compare the enhanced test images with the ground truth high-quality ones.
For this we use both the Point Signal-to-Noise Ratio (\emph{PSNR}) and the structural similarity index measure (\emph{SSIM})~\cite{SSIM}.
The former measures the amount of signal lost \wrt a reference signal (\eg, an image), 
the latter compares two images' similarity in terms of visually structured elements
and is known for its improved correlation with human perception, surpassing PSNR.

We adhere to the setup of~\cite{IKTVvG17} and train our model to map source photos to the domain of target DSLR images for each of three mobile cameras from the DPED dataset separately. Note that we use the DSLR photos in weak supervision only (without exploiting the pairwise correspondence between the source/target images): 
the adversarial discriminators are trained at each iteration with a random positive (\ie, DSLR) image and a random negative (\ie, non-DSLR) one.
For each mobile phone camera, we train two networks with different target images: first using the original DPED DSLR photos as target (noted "WESPE [DPED]"), second using the high-quality pictures from the DIV2K dataset~\cite{agustsson2017_ntire} (noted WESPE [DIV2K]).
Full-reference (PSNR, SSIM) scores calculated \wrt the DPED ground truth enhanced images are given in Table~\ref{tab:DPED_reference}.

\begin{figure*}[t!]
\centering
\setlength{\tabcolsep}{1pt}
\resizebox{0.9\linewidth}{!}
{
\begin{tabular}{cccc}
BlackBerry & BlackBerry & Sony & Sony \\
   \includegraphics[width=0.245\linewidth]{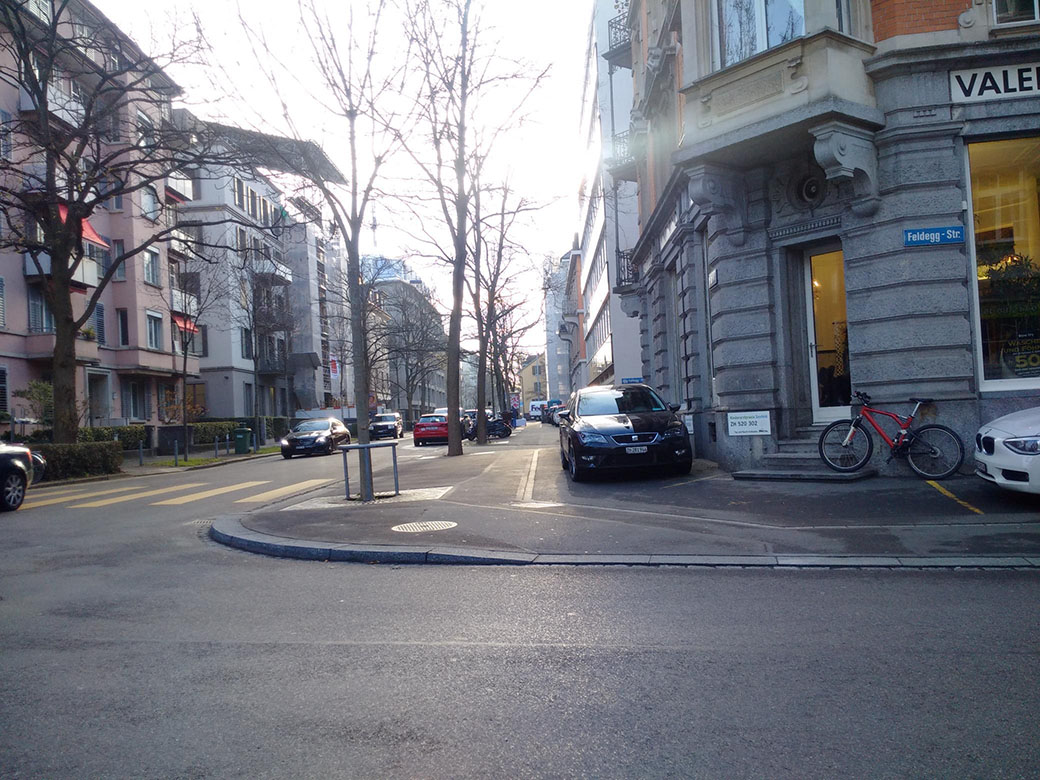}&
   \includegraphics[width=0.245\linewidth]{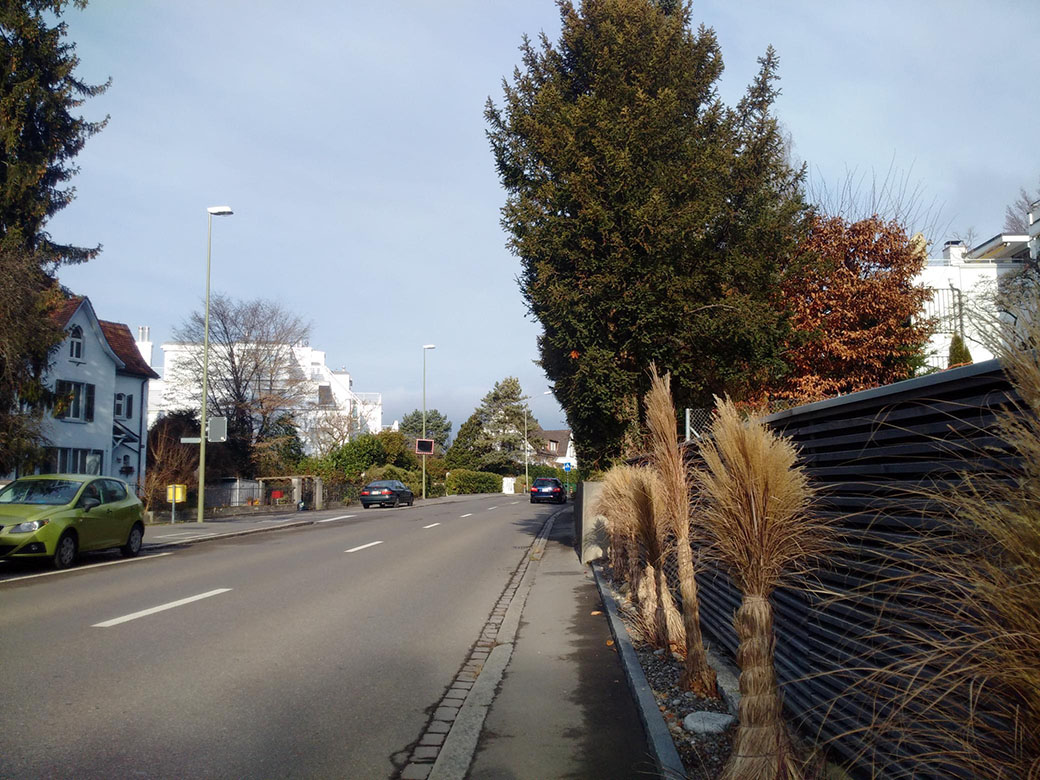}&
   \includegraphics[width=0.245\linewidth]{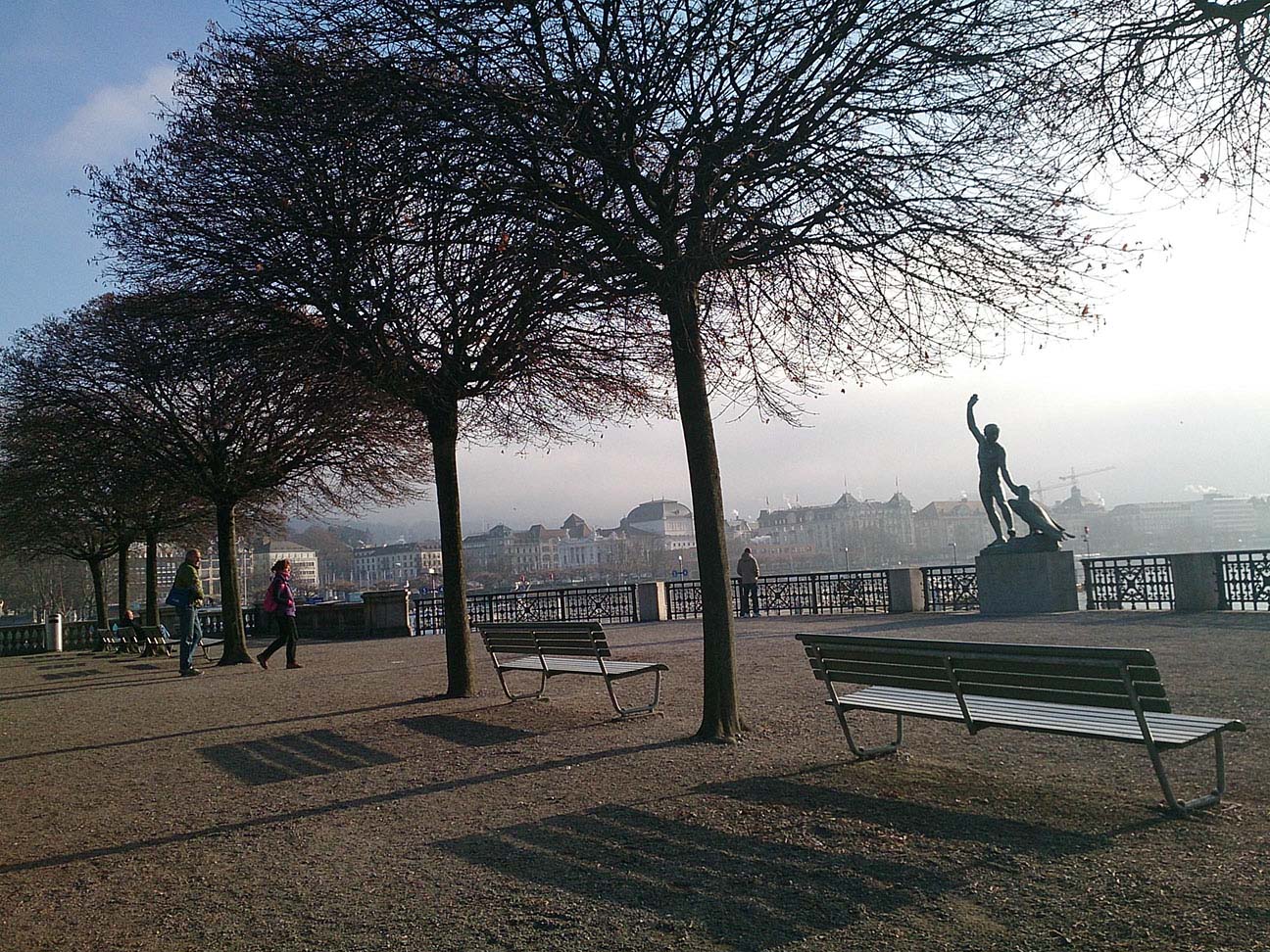}&
   \includegraphics[width=0.245\linewidth]{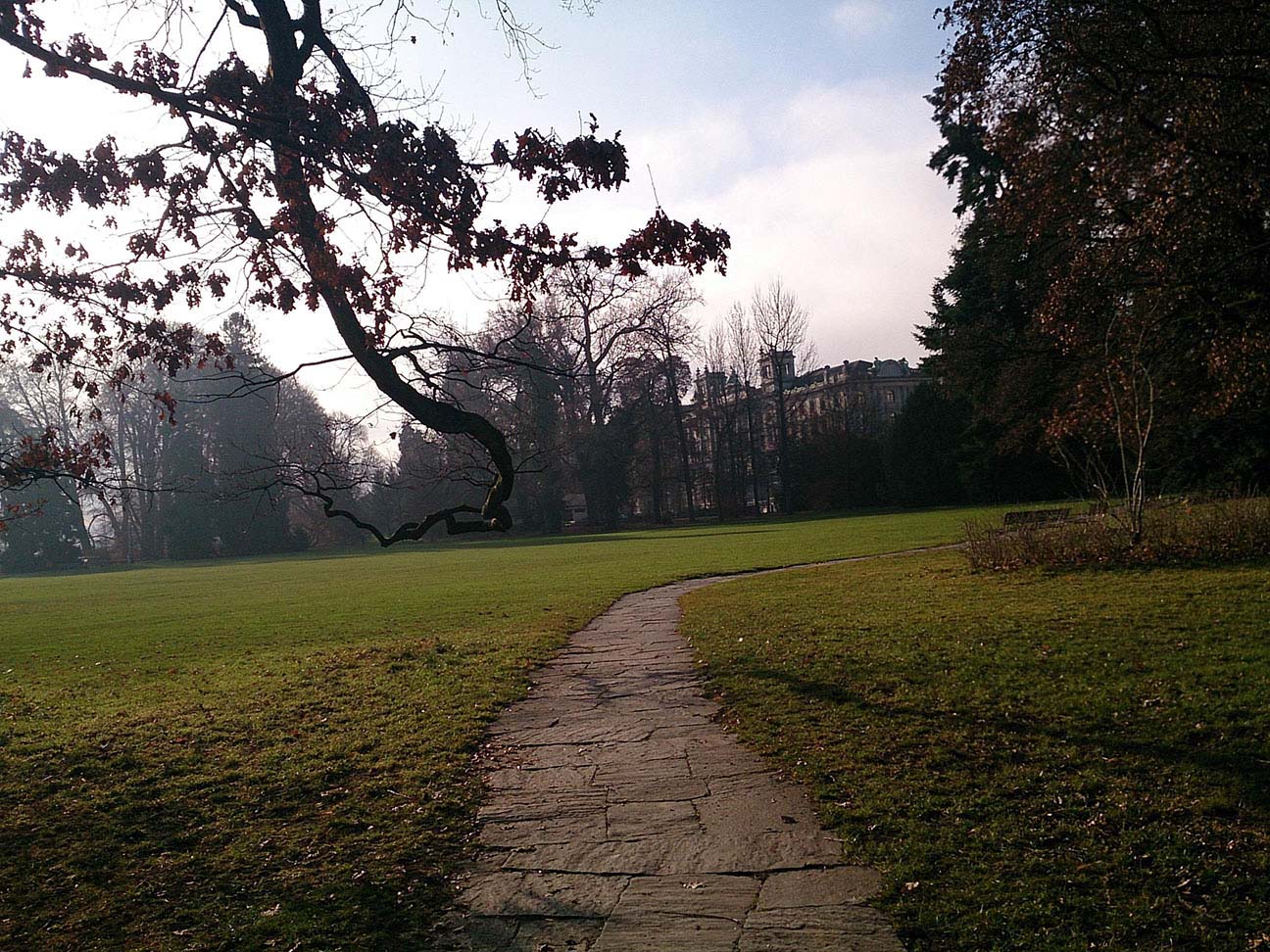}
   \\
   \includegraphics[width=0.245\linewidth]{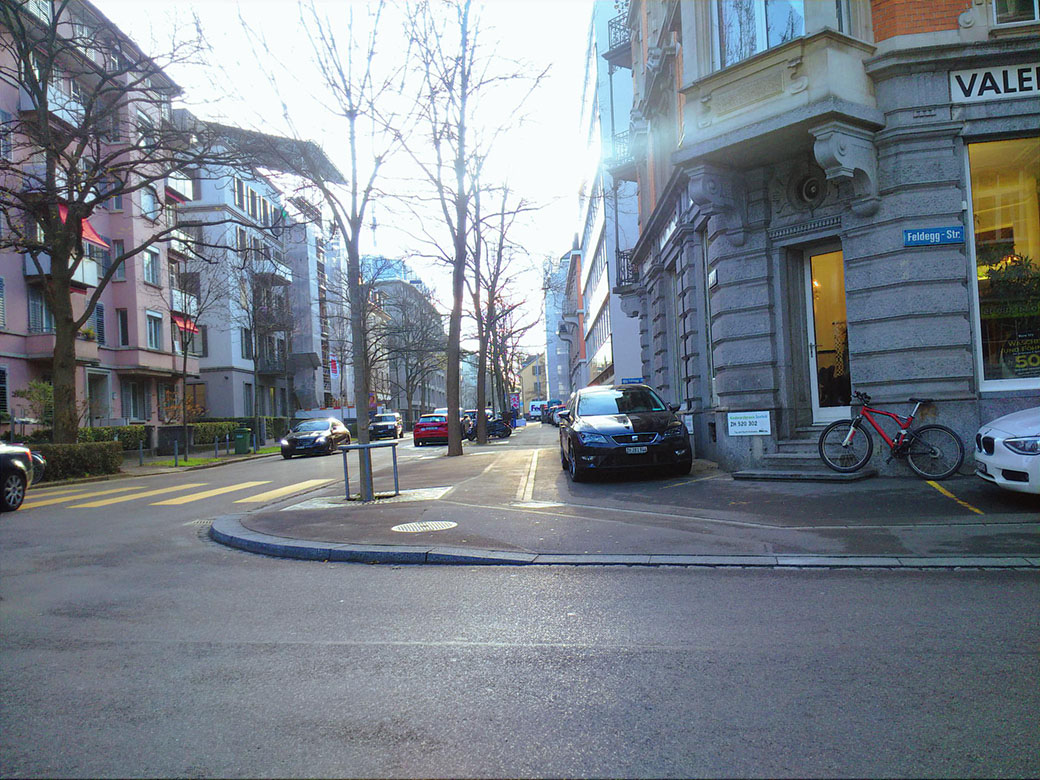}&
   \includegraphics[width=0.245\linewidth]{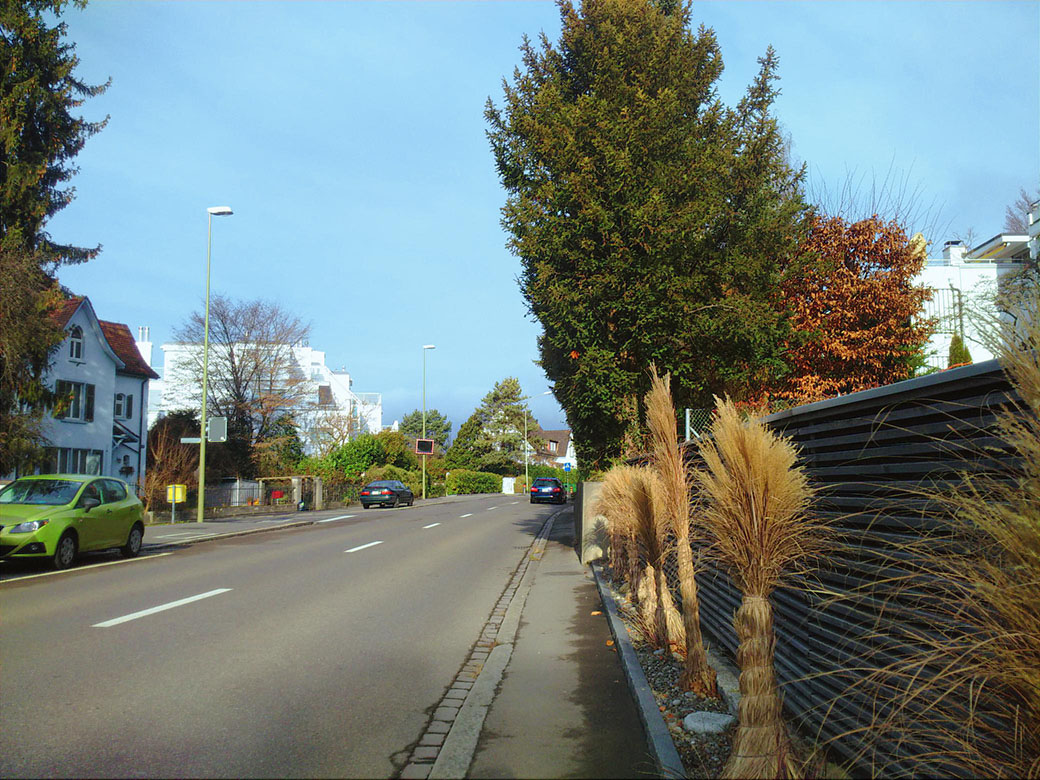}&
   \includegraphics[width=0.245\linewidth]{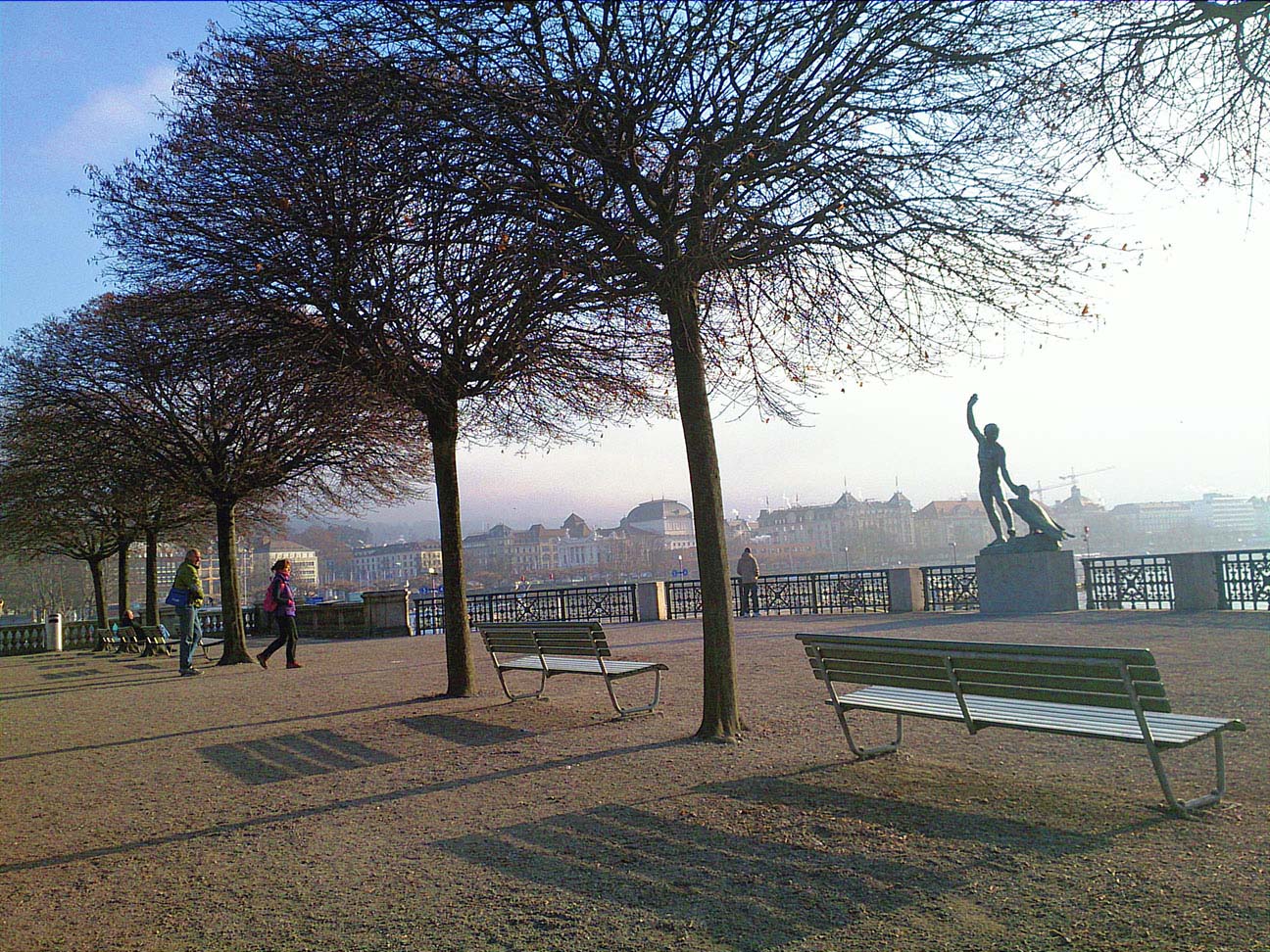}&
   \includegraphics[width=0.245\linewidth]{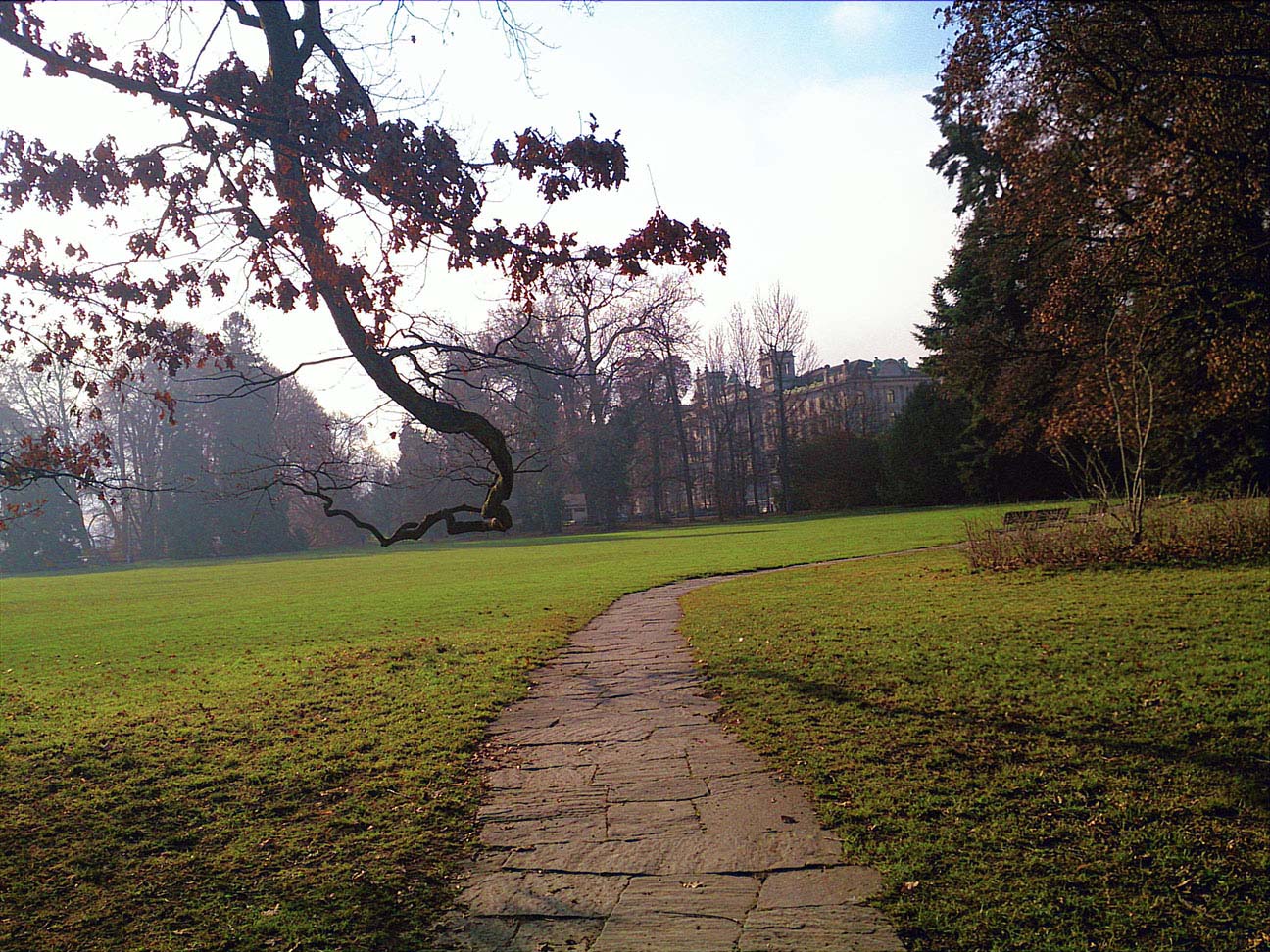} \\
\end{tabular}
}
 \caption{Original (top) vs. WESPE [DIV2K] enhanced (bottom) DPED images captured by BlackBerry and Sony cameras.}
\vspace{-2mm}
 \label{fig:enhancedexamples}
\end{figure*}

Our WESPE method trained with the DPED DSLR target performs better than the baseline method (APE).
Considering the better SSIM metric only, it is even almost as good as the network in~\cite{IKTVvG17} that uses a fully supervised approach and requires pixel-aligned ground truth images.
WESPE trained on DIV2K images as target (WESPE [DIV2K]) and tested \wrt DPED images degrades PSNR and SSIM scores compared to WESPE [DPED], but still remains above APE.
This is unsurprising as we are measuring proximity to known ground truth images laying in the domain of DPED DSLR photos (and not DIV2K): being close to it does not necessarily imply looking good.
Visually (see Figs.~\ref{fig:all_enhancements1} and~\ref{fig:enhancedexamples}), WESPE [DIV2K] seem to show crisper colors and we hypothesize they may be preferable, albeit further away from the ground truth image.
This also hints that using diverse data (DIV2K has a diverse set of sources) of high-quality images (\eg, with few noise) may be beneficial as well.
The following experiments try to confirm this.

\begin{table}[b!]
\centering
\caption{DPED dataset~\cite{IKTVvG17} with aligned images.}
\resizebox{\linewidth}{!}{
\begin{tabular}{l|rrrrr}
Camera source & Sensor & Image size & Photo quality & train images & test images \\
\hline
iPhone 3GS & 3MP & $2048\times 1536$ & Poor & 5614 & 113 \\
BlackBerry Passport & 13MP & $4160\times3120$ & Mediocre & 5902 & 113 \\
Sony Xperia Z & 13MP & $2592\times1944$ & Good & 4427 & 76 \\
Canon 70D DSLR &20MP & $3648\times2432$ & Excellent& 5902 & 113 \\
\end{tabular}
}
\label{tab:dped_dataset}
\end{table}

\subsection{No-reference evaluation in the wild}
\label{sec:experiments:wespe}

\begin{table}[b!]
\caption{Average PSNR and SSIM results on DPED test images. Best results are in \textbf{bold}.}
\centering
\resizebox{\linewidth}{!}
{
\begin{tabular}{l|cc|cc|cc|cc}
\multirow{3}{*}{DPED images} & \multicolumn{2}{c|}{\multirow{2}{*}{APE}} & \multicolumn{4}{c|}{Weakly Supervised}                                                   & \multicolumn{2}{c}{Fully Supervised}                  \\ 
& \multicolumn{2}{c|}{}                     & \multicolumn{2}{c|}{WESPE [DIV2K]} & \multicolumn{2}{c|}{WESPE [DPED]} & \multicolumn{2}{c}{\cite{IKTVvG17}} \\ 
                       & PSNR     & SSIM   & PSNR     & SSIM  & PSNR     & SSIM  & PSNR         & SSIM  \\ \hline
iPhone                 & 17.28    & 0.86   & 17.76    & 0.88  & 18.11    & 0.90  & \textbf{21.35}        & \textbf{0.92} \\ 
BlackBerry             & 18.91    & 0.89   & 16.71    & 0.91  & 16.78    & 0.91  & \textbf{20.66}        & \textbf{0.93} \\ 
Sony                   & 19.45    & 0.92   & 20.05    & 0.89  & 20.29    & 0.93  & \textbf{22.01}        & \textbf{0.94} \\ 
\end{tabular}
}
\label{tab:DPED_reference}
\end{table}

\begin{table*}[t]
\resizebox{\linewidth}{!}
{
\begin{tabular}{l|ccc|ccc|ccc|ccc|ccc}
\multirow{2}{*}{DPED images}&\multicolumn{3}{c|}{Original}&\multicolumn{3}{c}{APE}&\multicolumn{3}{c|}{\cite{IKTVvG17}}&\multicolumn{3}{c}{WESPE [DPED]}&\multicolumn{3}{c}{WESPE [DIV2K]}\\
   &entropy & bpp & CORNIA&entropy & bpp & CORNIA&entropy & bpp & CORNIA&entropy & bpp & CORNIA&entropy & bpp & CORNIA\\
\hline
iPhone & 7.29 & 10.67 & 30.85 & 7.40 & 9.33 & 43.65 & \textbf{7.55} & 10.94 & 32.35& 7.52 & 14.17 & 27.90& 7.52 & \textbf{15.13} & \textbf{27.40}\\
BlackBerry& 7.51 & 12.00 & 11.09& 7.55 & 10.19 & 23.19 & 7.51 & 11.39 & 20.62& 7.43 & 12.64 & 23.93 & \textbf{7.60} & \textbf{12.72} & \textbf{9.18}\\
Sony& 7.51 & 11.63 & 32.69 & \textbf{7.62} & 11.37 & 34.85& 7.53 & 10.90 & \textbf{30.54} & 7.59 & 12.05 & 34.77& 7.46 & \textbf{12.33} & 34.56\\
\end{tabular}
}
\caption{Average entropy, bit per pixel and CORNIA (lower is better) results on DPED test images. Best results are in \textbf{bold}.}
\vspace{-2mm}
\label{tab:noreference_dped_scores}
\end{table*}

\begin{figure*}[t!]
\centering
\setlength{\tabcolsep}{1pt}
\resizebox{0.9\linewidth}{!}
{
\begin{tabular}{cccc}
    Cityscapes & Cityscapes & KITTI & KITTI \\
   \includegraphics[width=0.245\linewidth]{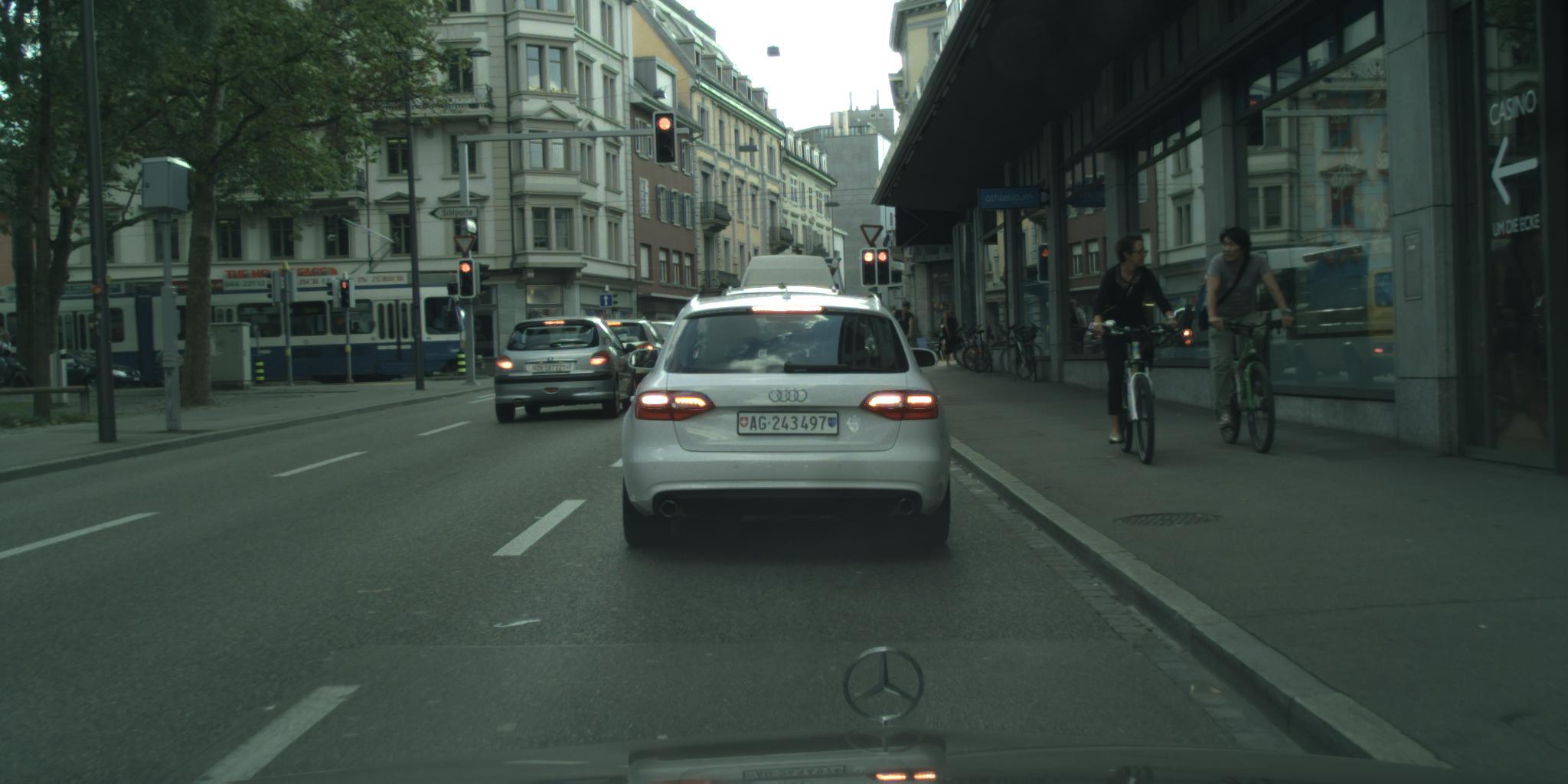}&
   \includegraphics[width=0.245\linewidth]{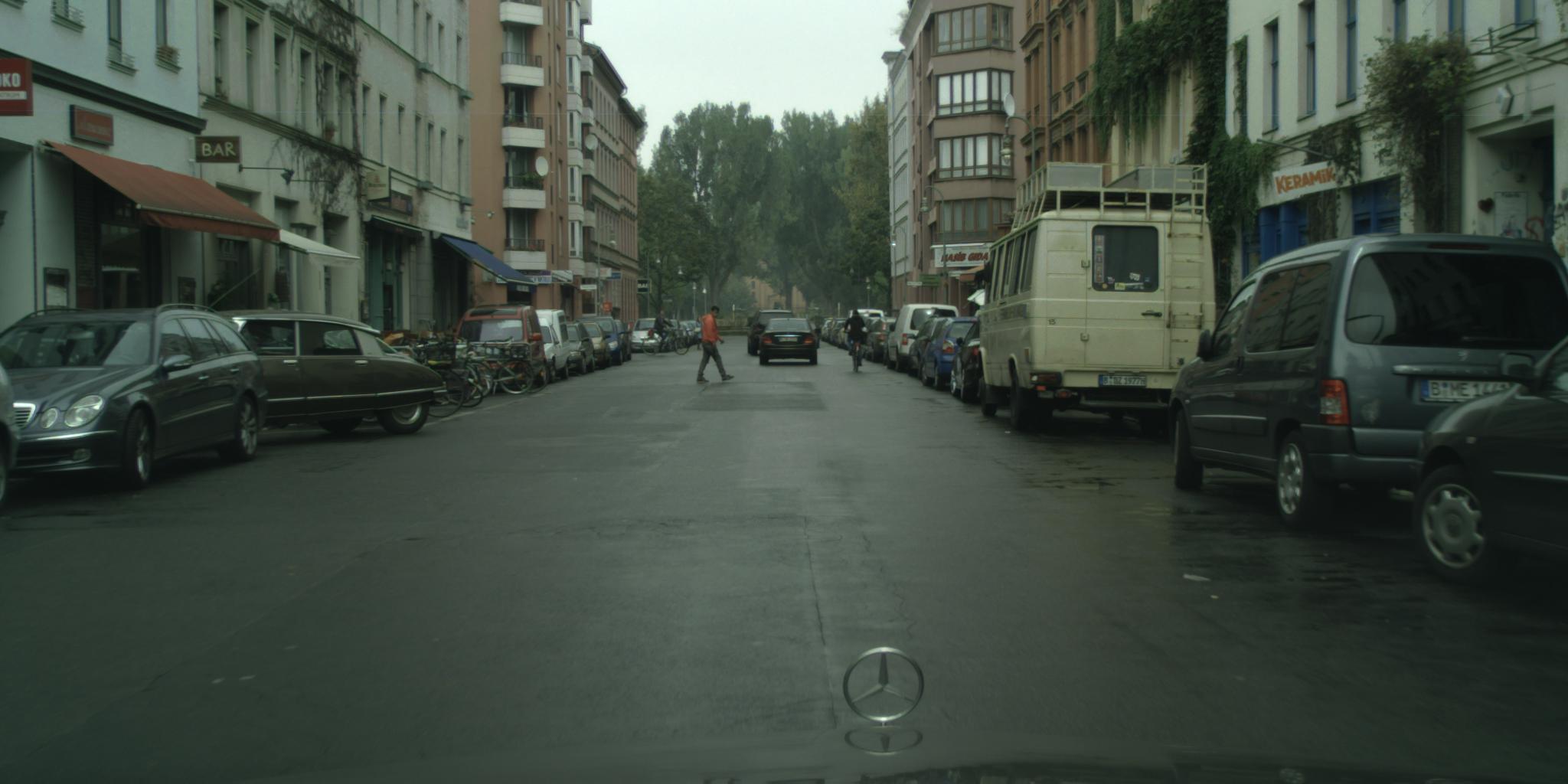}&
   \includegraphics[width=0.245\linewidth]{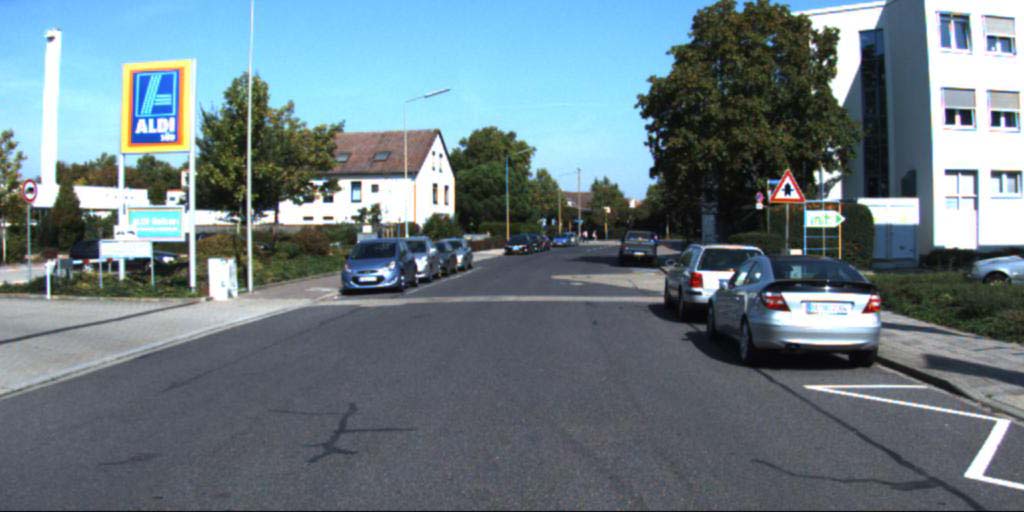}&
   \includegraphics[width=0.245\linewidth]{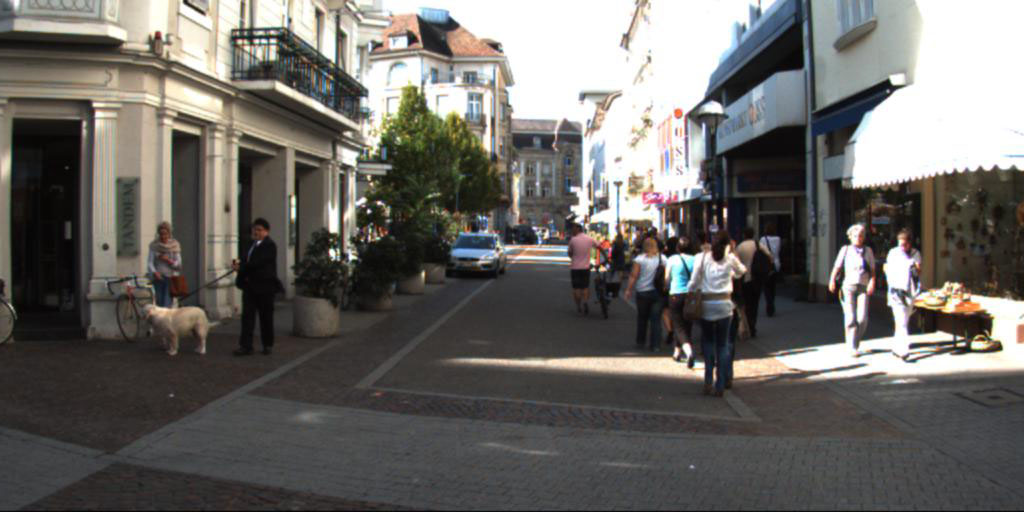}\\
   \includegraphics[width=0.245\linewidth]{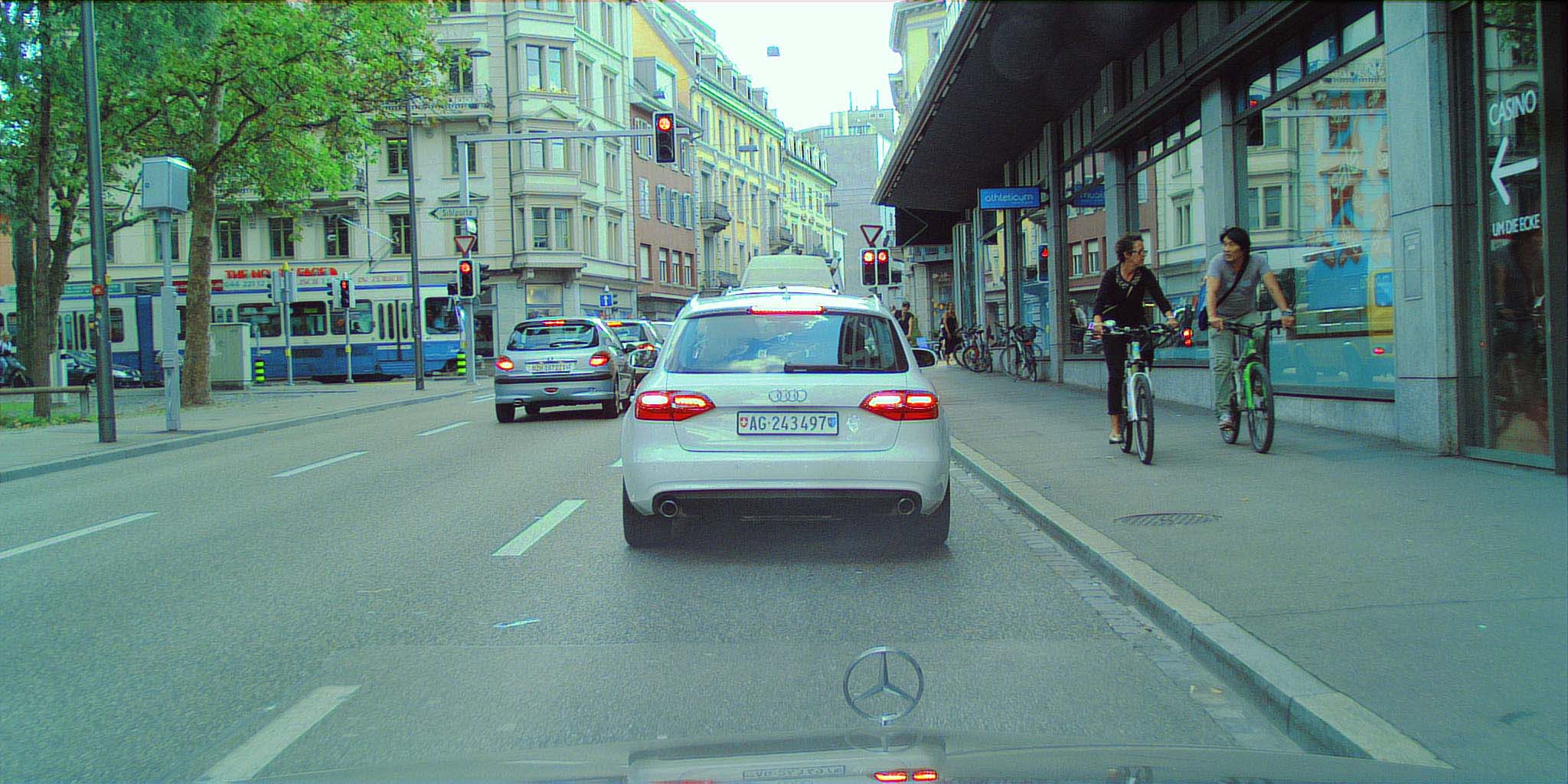}&
   \includegraphics[width=0.245\linewidth]{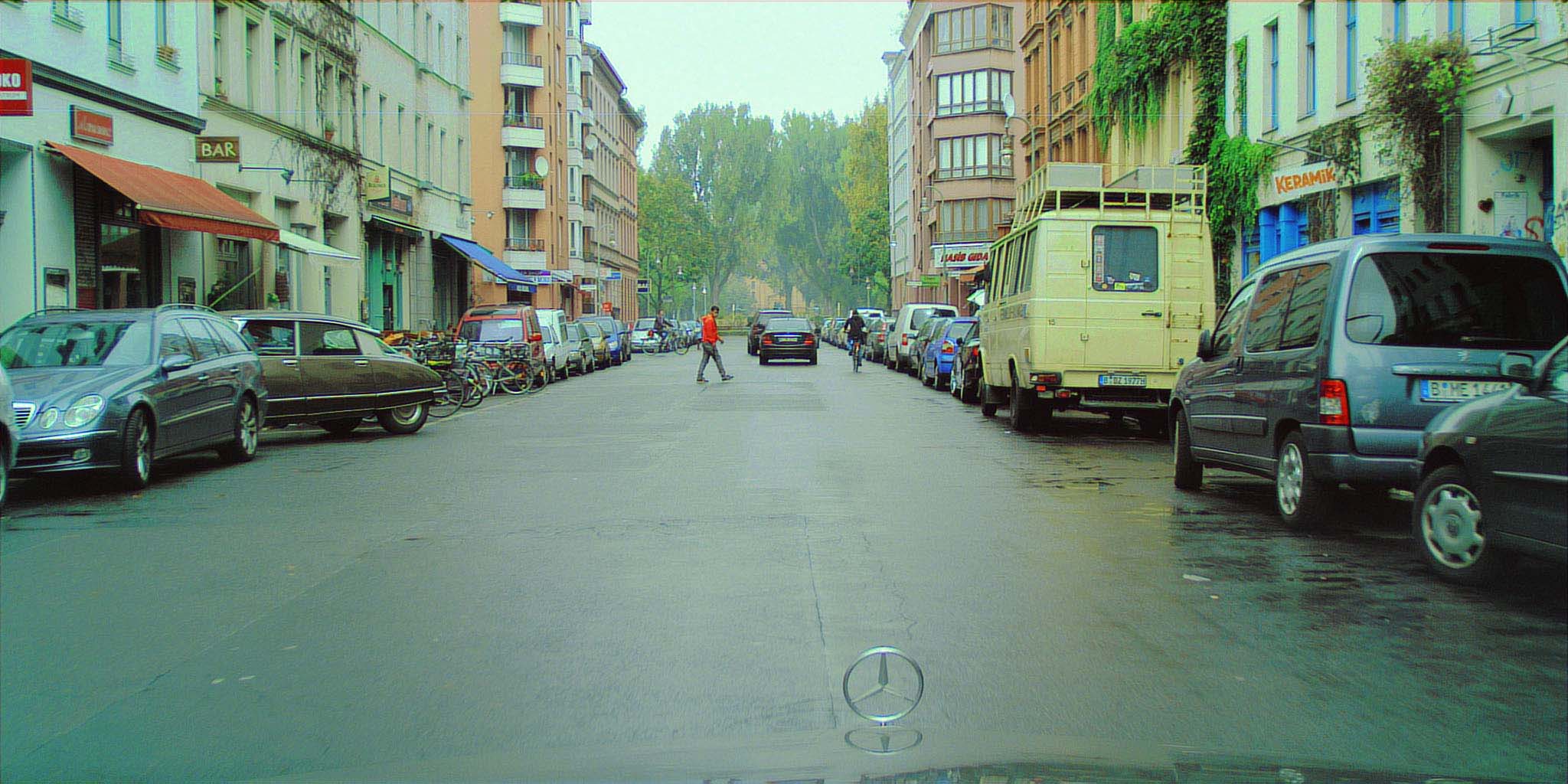}&
   \includegraphics[width=0.245\linewidth]{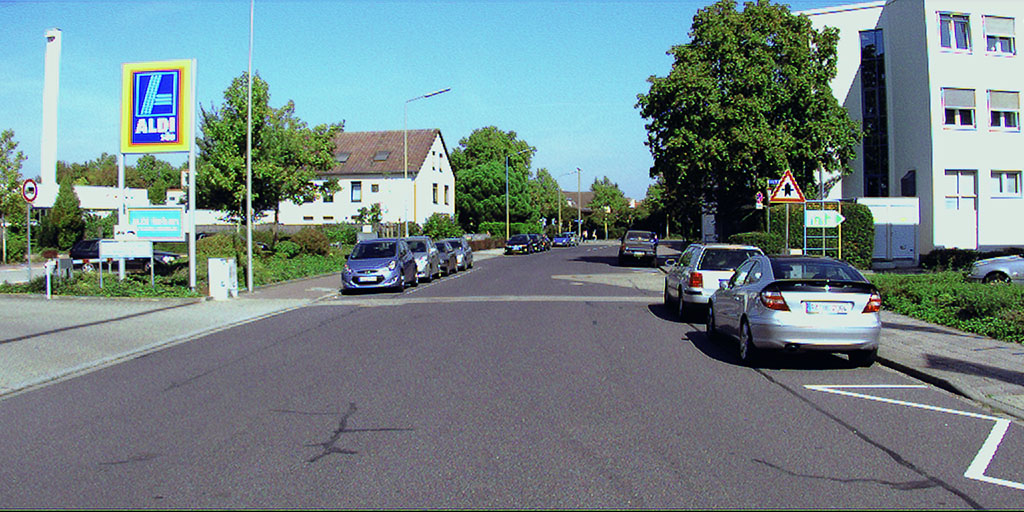}&
   \includegraphics[width=0.245\linewidth]{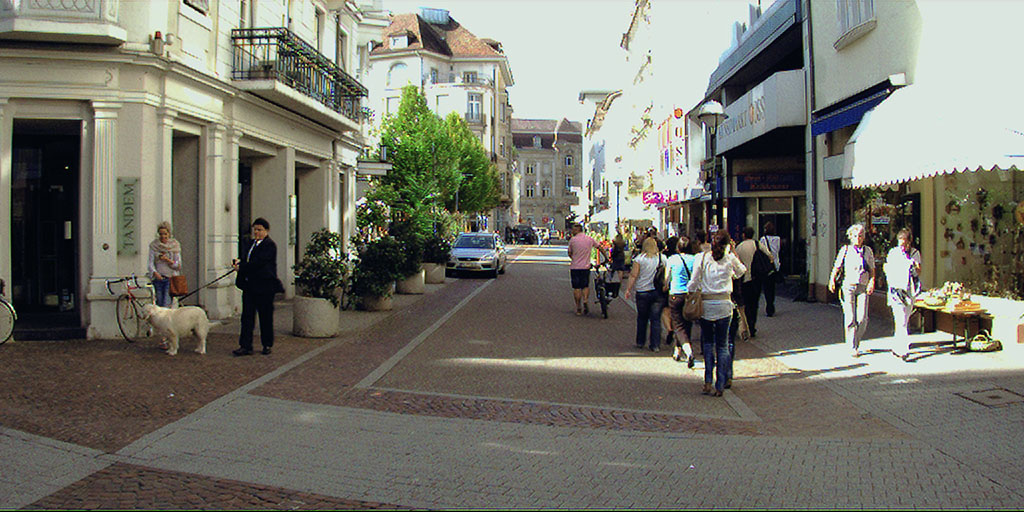} \\
\end{tabular}
}
 \caption{Examples of original (top) vs. enhanced (bottom) images for the Cityscapes and KITTI datasets.}
 \label{fig:enhanced_cityscape}
\end{figure*}

WESPE does not require before/after ground truth correspondences to be trained, so in this section we train it on various datasets in the wild whose main characteristics are shown in Table~\ref{tab:in_the_wild_datasets} as used in our experiments.
Besides computing no-reference scores for the results obtained in the previous section, we complement the DPED dataset containing photos from older phones with pictures taken by phones marketed as having state-of-the-art cameras: the iPhone~6, HTC One~M9 and \mbox{Huawei~P9}.
To avoid compression artifacts which may occur in online-crawled images, we did a manual collection in a peri-urban environment of thousands of pictures for each phone/camera.
We additionally consider two widely-used datasets in Computer Vision and learning: the Cityscapes~\cite{cityscapes} and KITTI~\cite{kitti} public datasets.
They contain a large-scale set of urban images of low quality, which forms a good use case for automated quality enhancement.
That is, Cityscapes contains photos taken by a dash-camera (it lacks image details, resolution and brightness), while KITTI photos are brighter, but only half the resolution, disallowing sharp details (see Figure~\ref{fig:enhanced_cityscape}).
Finally, we use the recent DIV2K dataset~\cite{agustsson2017_ntire} of high quality images and diverse contents and camera sources as a target for our WESPE training.

\begin{table}
\centering
\resizebox{\linewidth}{!}
{
\begin{tabular}{l|rrrrr}
Camera source & Sensor & Image size & Photo quality & train images & test images \\
\hline
KITTI~\cite{kitti}& N/A & $1392\times 512$ & Poor & 8458 & 124 \\
Cityscapes~\cite{cityscapes}& N/A & $2048\times 1024$ & Poor & 2876 & 143 \\
HTC One M9 & 20MP & $5376\times 3752$ & Good & 1443 & 57\\
Huawei P9 & 12MP & $3968\times 2976$ & Good & 1386 & 57 \\
iPhone 6 & 8MP & $3264\times 2448$ & Good & 4011 & 57\\
Flickr Faves Score (FFS) & N/A & $>1600\times1200$ & Poor-to-Excellent& 15600 & 400 \\
DIV2K~\cite{agustsson2017_ntire} & N/A & $\sim2040\times1500$ & Excellent& 900 & 0\\
\end{tabular}
}
\caption{Datasets in the wild as used in our experiments. No aligned image pairs from different cameras are available.}
\vspace{-4mm}
\label{tab:in_the_wild_datasets}
\end{table}

Importantly, here we evaluate all images with no-reference quality metrics, that will give an absolute image quality score, not a proximity to a reference.
For objective quality measurement, we mainly focus on the Codebook Representation for No-Reference Image Assessment (\emph{CORNIA})~\cite{CORNIA}: it is a perceptual measure mapping to average human quality assessments for images.
Additionally, we compute typical signal processing measures, namely image \emph{entropy} (based on pixel level observations) and bits per pixel (\emph{bpp}) of the PNG lossless image compression.
Both image entropy and bpp are indicators of the quantity of information in an image.
We train WESPE to map from one of the datasets mentioned above to the DIV2K image dataset as target.
We also report absolute quality measures (\ie, bbp, entropy and CORNIA scores) on original DPED images as well as APE-enhanced, \cite{IKTVvG17}-enhanced and WESPE-enhanced ([DPED] and [DIV2K] variants) images in Table~\ref{tab:noreference_dped_scores}, and take the best-performing methods to compare on the remaining datasets in Table~\ref{tab:cornia}.

Table~\ref{tab:noreference_dped_scores} shows that the DIV2K variant of WESPE generates the best overall image quality, surpassing~\cite{IKTVvG17} and the WESPE variant that targets DPED DSLR images.
This confirms the impression that proximity to ground truth is not the only matter of importance.
This table also shows that improvement is stronger for low-quality camera's (iPhone and Blackberry) than for the better Sony camera, which probably benefits less from the WESPE image healing.
Moreover, targeting the DIV2K image quality domain seems to improve over the DPED DSLR domain: WESPE [DIV2K] generally improves or competes with WESPE [DPED] and even the fully supervised~\cite{IKTVvG17} network.

\begin{figure*}[t!]
\setlength{\tabcolsep}{1pt}
\centering
\resizebox{0.9\linewidth}{!}
{
\begin{tabular}{cccc}
   \includegraphics[width=0.33\linewidth]{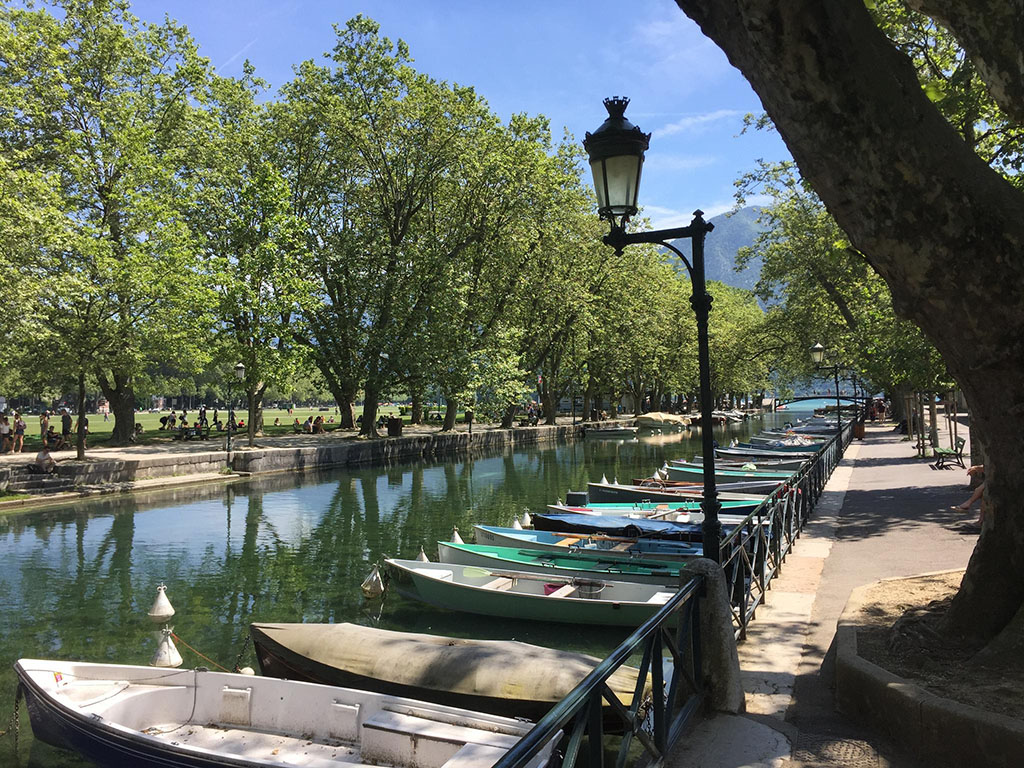}&
   \includegraphics[width=0.33\linewidth]{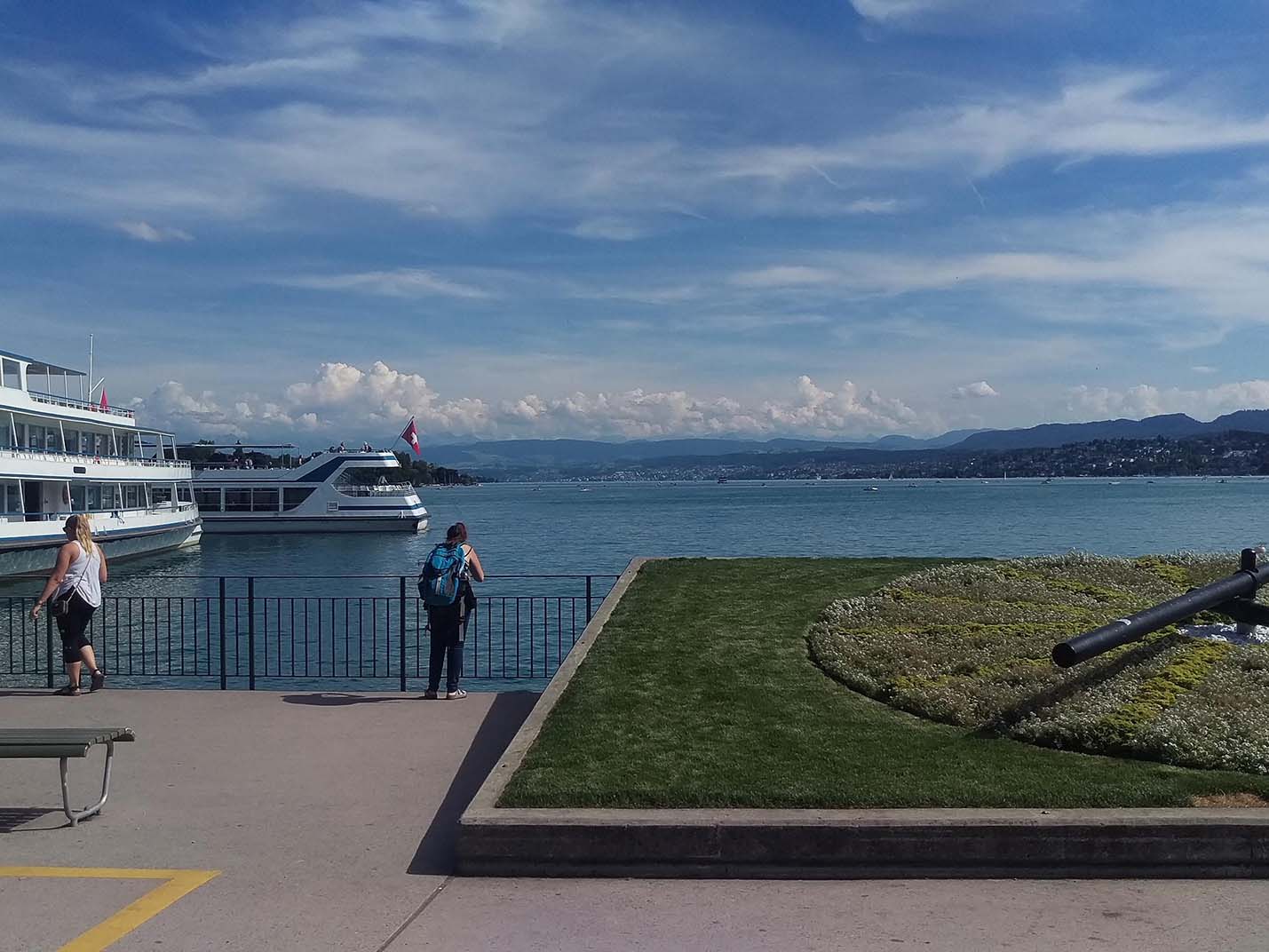}&
   \includegraphics[width=0.33\linewidth]{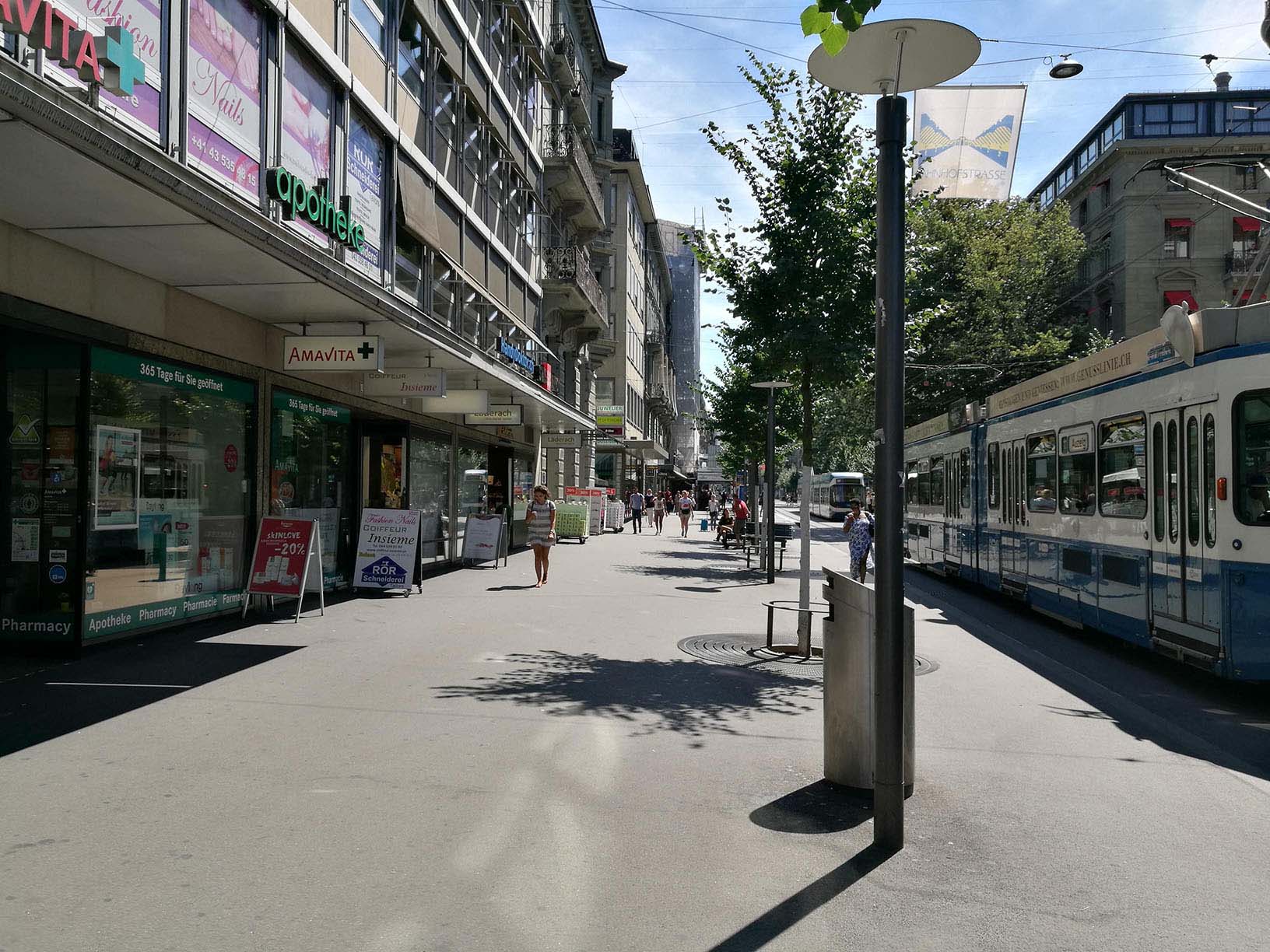}&
   \\
   \includegraphics[width=0.33\linewidth]{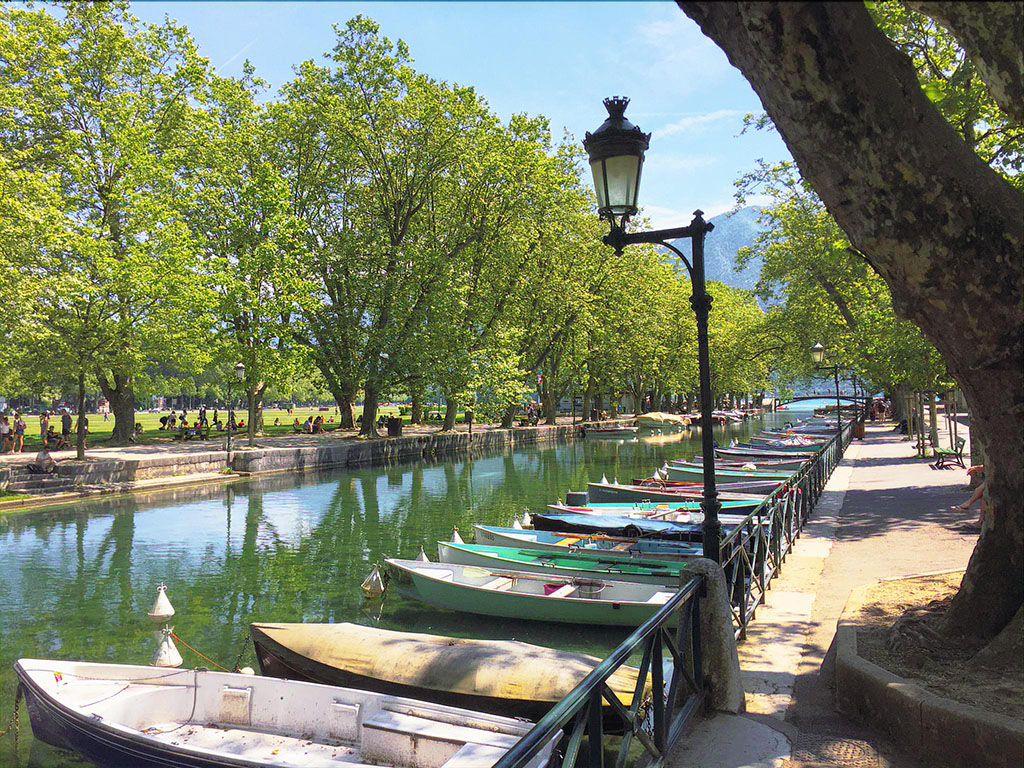}&
   \includegraphics[width=0.33\linewidth]{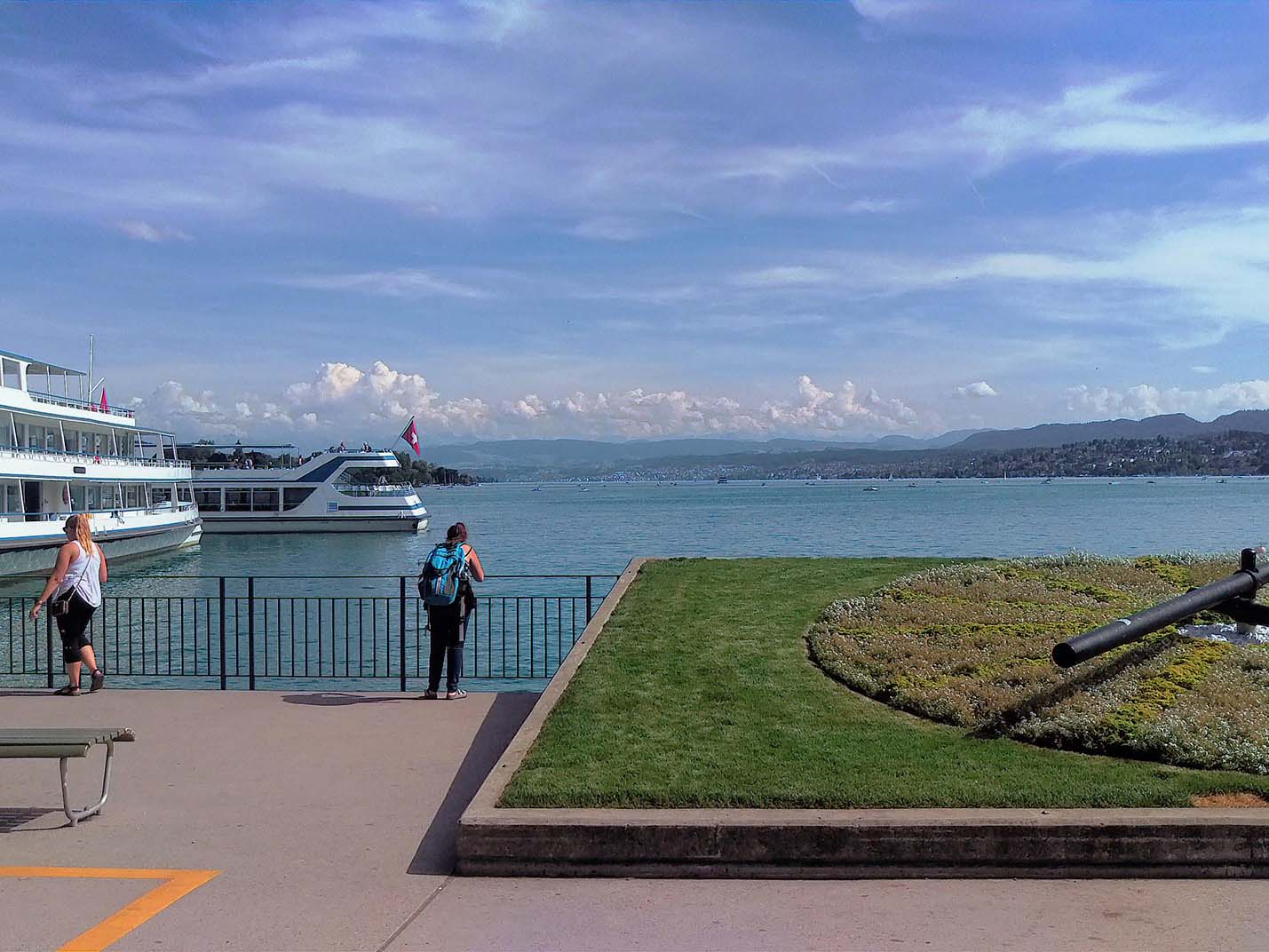}&
   \includegraphics[width=0.33\linewidth]{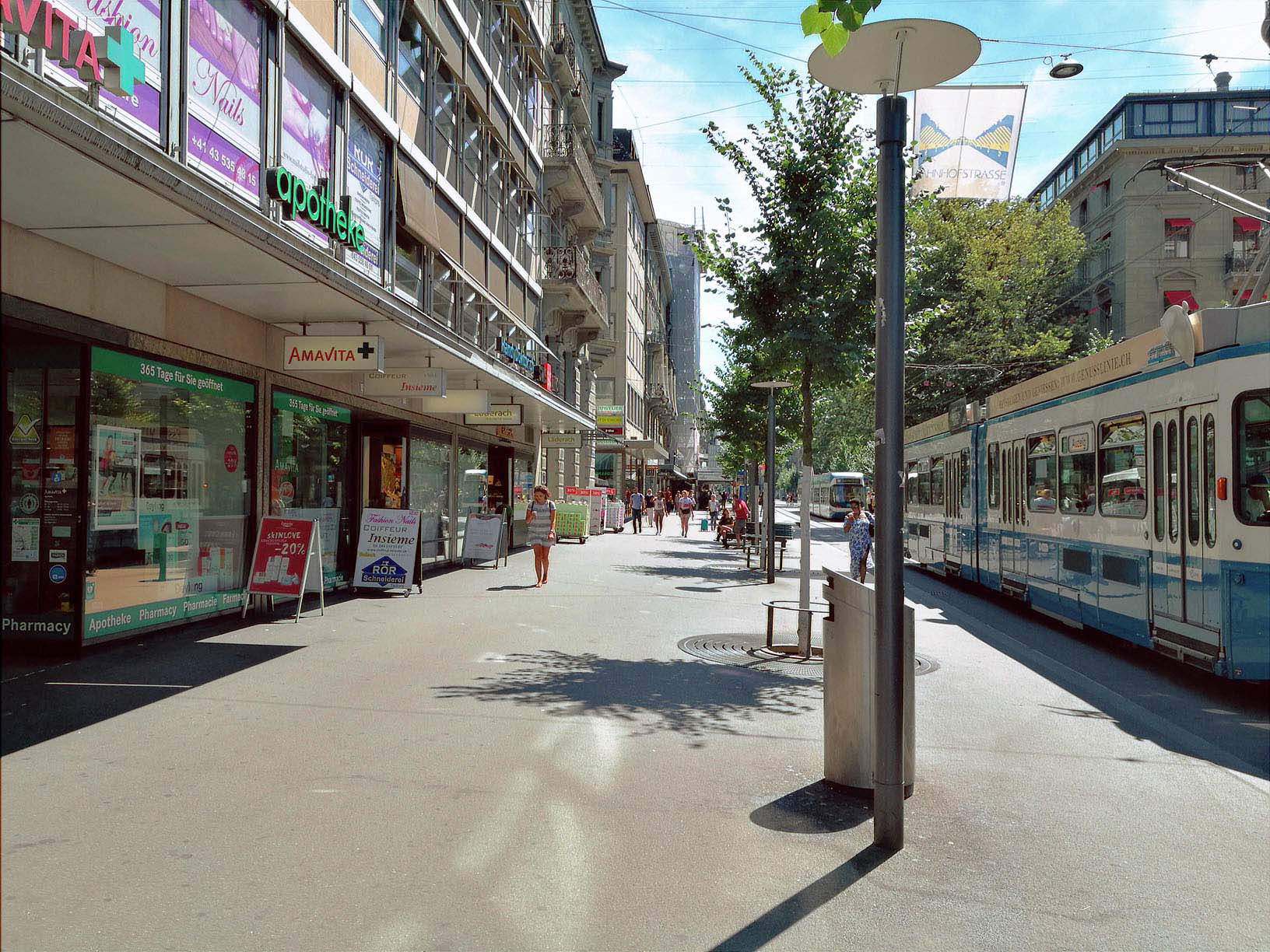}&
\end{tabular}
}
 \caption{Original (top) vs. enhanced (bottom) images for iPhone~6, HTC~One~M9 and Huawei~P9 cameras.}
\vspace{-2mm}
 \label{fig:enhanced_huawei}
\end{figure*}

On datasets in the wild (Table~\ref{tab:cornia}), WESPE and APE improve the original images on all metrics on the urban images (KITTI and Cityscapes). WESPE demonstrates significantly better results on the CORNIA and bpp metrics, but also on image entropy. Recall that KITTI and Cityscapes consist of images of poor quality, and our method is successful in healing such pictures.
On the smartphones, whose pictures are already of high quality, our method shows improved bpp and slightly worse CORNIA scores, while keeping image entropy on par.
The latter findings are quite ambiguous, since visual results for the urban (Figure~\ref{fig:enhanced_cityscape}) and phone datasets (Figure~\ref{fig:enhanced_huawei}) demonstrate that there is a significant image quality difference that is not fully reflected by the entropy, bpp, and CORNIA quantitative numbers as proxy measures for perceived image quality. Moreover, since the correlation between objective scores and human perception can be debatable, in the following subsections we provide a complementary subjective quality evaluation.

\subsection{User study}
\label{sec:experiments:userstudy}
Since the final aim is to improve both the quality and aesthetics of an input image, we conducted a user study comparing subjective evaluation of the original, APE-enhanced and WESPE-enhanced photos with DIV2K as target, for the 5 datasets in the wild (see section~\ref{sec:experiments:wespe} and Table~\ref{tab:in_the_wild_datasets}).
To assess subjective quality, we chose a pairwise forced choice method. The user's task was to choose the preferred picture among two displayed side by side. No additional selection criteria were specified, and users were allowed to zoom in and out at will without time restriction.
Seven pictures were randomly taken from the test images of each dataset (\ie, 35 pictures total).
For each image, the users were shown a before vs. after WESPE-enhancement pair and a APE-enhanced vs. WESPE-enhanced pair to compare.
38 people participated in this survey and fulfilled the $35\times 2$ selections.
The question sequence, as well as the sequence of pictures in each pair were randomized for each user.
Preference proportions for each choice are shown in Table~\ref{tab:user-study-results}.

\begin{table}[b!]
\vspace{-2mm}
\centering
\resizebox{\linewidth}{!}
{
\begin{tabular}{l|ccccc}
Setting           & {\small Cityscapes} & {\small KITTI} & {\small HTC M9} & {\small Huawei P9}  & {\small iPhone 6} \\ \hline
WESPE vs Original & 0.94$\pm$0.03 & 0.81$\pm$0.10 & 0.73$\pm$0.08 & 0.63$\pm$0.11 & 0.70$\pm$0.10\\
WESPE vs APE 	  & 0.96$\pm$0.03 & 0.65$\pm$0.16 & 0.53$\pm$0.09 & 0.44$\pm$0.12 & 0.62$\pm$0.15 \\
\end{tabular}
}
\caption{User study results. The fraction of times WESPE result was preferred over original or APE-enhanced images.
}
\label{tab:user-study-results}
\end{table}

\begin{table*}[t!]
\centering
\caption{Average entropy, bit per pixel and CORNIA scores on five test datasets. Best results are in \textbf{bold}.}
\resizebox{0.7\linewidth}{!}
{
\begin{tabular}{l|ccc|ccc|ccc}
\multirow{2}{*}{Images}&\multicolumn{3}{c|}{Original}&\multicolumn{3}{c|}{APE}&\multicolumn{3}{c}{WESPE [DIV2K]}\\
   &entropy & bpp & CORNIA&entropy & bpp & CORNIA&entropy & bpp & CORNIA\\
\hline
Cityscapes & 6.73 & 8.44 & 43.42 &7.30 & 6.74 & 46.73  & \textbf{7.56} & \textbf{11.59} & \textbf{32.53} \\
KITTI & 7.12 & 7.76 & 55.69 & \textbf{7.58} & 10.21 & \textbf{37.64}& 7.55 & \textbf{11.88} & 39.09 \\
\hline
HTC One M9 & 7.51 & 9.52 & \textbf{23.31}  & 7.64 & 9.64 & 28.46 & \textbf{7.69} & \textbf{12.99} & 26.35\\
Huawei P9 & 7.71 & 10.60 & \textbf{20.63} & \textbf{7.78} & 10.27 & 25.85 & 7.70 & \textbf{12.61} & 27.52 \\
iPhone 6 & 7.56 & 11.65 & \textbf{24.67}  & \textbf{7.57} & 9.25 & 35.82 & 7.53 & \textbf{13.44} & 28.51\\
\end{tabular}
}
\label{tab:cornia}
\vspace{-2mm}
\end{table*}

WESPE-improved images are on average preferred over non-enhanced original images, even by a vast majority in the case of Cityscapes and KITTI datasets.
On these two, the WESPE results are clearly preferred over the APE ones, especially on the Cityscapes dataset.
On the modern phone cameras, users found it difficult to distinguish the quality of the WESPE-improved and APE-improved images, especially when the originals were already of good quality, on the HTC~One~M9 or Huawei~P9 cameras in particular.

\subsection{Flickr Faves Score}
\label{sec:experiments:ffs}

Gathering human-perceived photo quality scores is a tedious hence non-scalable process. To complement this, we train a virtual rater to mimic Flickr user behavior when adding an image to their favorites.
Under the assumption that users tend to add better rather than lower quality images to their Faves, we train a binary classifier CNN to predict favorite status of an image by an average user, which we call the Flickr Faves Score (FFS).

First, we collect a \textit{Flickr Faves Score dataset} (FFSD) consisting of 16K photos randomly crawled from Flickr along with their number of views and Faves. Only images of resolution higher than $1600\times1200$ pixels were considered and then cropped and resized to HD-resolution.
We define the FFS score of an image as the number of times is was fav'ed over the number of times it was viewed ($FFS(I)=\#F(I)/\#V(I)$), and assume this strongly depends on overall image quality.
We then binary-label all images as either low --or high-quality based the median FFS: below median is low-quality, above is high-quality.
This naive methodology worked fine for our experiments (see results below): we leave analyzing and improving it for future work.

\begin{table}[t]
 \caption{FFS scores on the DPED dataset.}
 \vspace{-2mm}
\centering
\resizebox{\linewidth}{!}{
\begin{tabular}{l|cccc}
&&fully&\multicolumn{2}{c}{Weakly Supervised}\\
\multirow{2}{*}{DPED images} & \multirow{2}{*}{original} & Supervised & WESPE [DPED] & \textbf{WESPE [DIV2K]} \\
 & & \cite{IKTVvG17} & (ours) & \textbf{(ours)} \\
\hline
iPhone & 0.3190 & 0.5093 & 0.5341 & \textbf{0.6155} \\
Blackberry & 0.4765 & 0.5366 & 0.5904 & \textbf{0.6001} \\
Sony & 0.5694 & 0.6572 & 0.6774 & \textbf{0.6828} \\
\hline
average& 0.4550& 0.5677 & 0.6006 & \textbf{0.6328}\\
\end{tabular}
}
\label{tab:oracle_dped}
\end{table}

Next, we train a VGG19-style~\cite{simonyan2014very} CNN on random $224\times 224$px patches to classify image Fave status and achieve $68.75\%$ accuracy on test images.
The network was initialized with VGG19 weights pre-trained on ImageNet~\cite{russakovsky2015imagenet}, and trained until the early stopping criterion is met with a learning rate of 5e-5 and a batch size of 25.
We split the data into training, validation and testing subsets of 15.2K, 400 and 400 images, respectively.
Note that using HD-resolution inputs would be computationally infeasible while downscaling would remove image details and artifacts important for quality assessment.
We used a single patch per image as more did not increase the performance.

We use this CNN to label both original and enhanced images from all datasets mentioned in this paper as Fave or not.
In practice, we do this by averaging the results for five unique crops from each image (the identical crops are used for both original and enhanced photos).
Per-dataset average FFS scores are shown in Tables~\ref{tab:oracle_dped} and~\ref{tab:oracle_wespe}.
Note that this labeling differs from pairwise preference selection as in our user study of section~\ref{sec:experiments:userstudy}: it's an absolute rating of images in the wild, as opposed to a limited pairwise comparison.

Our first observation is that the FFS scorer behaves coherently with all observations about DPED:
the three smartphones' original photos that were termed as `poor', `mediocre' and `average' in~\cite{IKTVvG17} have according FFS scores (Table~\ref{tab:oracle_dped}, first column), and the more modern cameras have FFS scores that are similar to the best DPED smartphone (\ie, Sony) camera (Table~\ref{tab:oracle_wespe}, first column).
Finally, poorer-quality images in the Cityscapes and KITTI datasets score significantly lower.
Having validated our scalable virtual FFS rater, one can note in Tables~\ref{tab:oracle_dped} and~\ref{tab:oracle_wespe} that the FFS scores of WESPE consistently indicate improved quality over original images or the ones enhanced by the fully supervised method of~\cite{IKTVvG17}.
Furthermore, this confirms our (now recurrent) finding that the [DIV2K] variant of WESPE improves over the [DPED] one.

\begin{table}[t]
\caption{FFS scores on five test datasets in the wild.}
\vspace{-2.6mm}
\centering
\resizebox{0.7\linewidth}{!}{
\begin{tabular}{l|cc}
Images & Original & \textbf{WESPE [DIV2K]}\\
\hline
Cityscapes & 0.4075 & \textbf{0.4339} \\
KITTI & 0.3792 & \textbf{0.5415} \\
\hline
HTC One M9& 0.5194 & \textbf{0.6193} \\
Huawei P9& 0.5322 & \textbf{0.5705} \\
iPhone 6& 0.5516 & \textbf{0.7412} \\
\hline
Average& 0.4780 & \textbf{0.5813}\\
\end{tabular}
}
\label{tab:oracle_wespe}
\vspace{-0mm}
\end{table}

\section{Conclusion}
\label{sec:conclusion}

In this work, we presented WESPE~-- a weakly supervised solution for the image quality enhancement problem. In contrast to previously proposed approaches that required strong supervision in the form of aligned source-target training image pairs, this method is free of this limitation. That is, it is trained to map low-quality photos into the domain of high-quality photos without requiring any correspondence between them: only two separate photo collections representing these domains are needed. To solve the problem, we proposed a transitive architecture that is based on GANs and loss functions designed for accurate image quality assessment. The method was validated on several publicly available datasets with different camera types. Our experiments reveal that WESPE demonstrates the performance comparable or surpassing the traditional enhancers and competes with the current state of the art supervised methods, while relaxing the need of supervision thus avoiding tedious creation of pixel-aligned datasets.

\bibliographystyle{ieee}

\end{document}